\documentclass[acmtog]{acmart}
\acmSubmissionID{313}

\usepackage{booktabs} %

\usepackage[ruled]{algorithm2e} %

\SetAlFnt{\small}
\SetAlCapFnt{\small}
\SetAlCapNameFnt{\small}
\SetAlCapHSkip{0pt}

\usepackage{graphicx}
\usepackage{amsmath}
\usepackage{booktabs}
\usepackage{multirow}
\usepackage{colortbl}
\usepackage{makecell}
\usepackage{diagbox}
\usepackage{svg}
\usepackage{pgfplots}
\pgfplotsset{
    compat=1.18,
    every axis plot/.append style={thick},
}

\usepackage{xcolor}
\definecolor{myred}{rgb}{0.8,0,0}
\definecolor{mygreen}{rgb}{0,0.8,0}
\definecolor{myblue}{rgb}{0.0,0.0,0.0}
\usepackage{pifont}%

\usepackage[normalem]{ulem} %

\newcommand{\equref}[1]{Eq.~(\ref{#1})}

\newcommand{\eg}{\textit{e.g}.}
\newcommand{\ie}{\textit{i.e}.}

\acmJournal{TOG}

\setcopyright{acmlicensed}
\acmJournal{TOG}
\acmYear{2023} \acmVolume{42} \acmNumber{6} \acmArticle{273} \acmMonth{12} \acmPrice{15.00}\acmDOI{10.1145/3618336}

\begin{document}
\title{Reconstructing Close Human Interactions from Multiple Views}

\author{Qing Shuai}
\orcid{0000-0002-2096-7599}
\authornote{The first two authors contributed equally to this work.}
\email{s_q@zju.edu.cn}
\affiliation{
 \institution{State Key Laboratory of CAD\&CG, Zhejiang University}
 \city{Hangzhou}
 \country{China}
}

\author{Zhiyuan Yu}
\orcid{0000-0001-7220-7789}
\authornotemark[1]
\email{zyuaq@ust.hk}
\affiliation{
 \institution{Hong Kong University of Science and Technology}
 \department{Mathematics}
 \city{Hong Kong}
 \country{China}
 }

\author{Zhize Zhou}
\orcid{0000-0001-9576-1686}
\email{zhouzhize@cupes.edu.cn}
\affiliation{
 \institution{Capital University of Physical Education and Sports}
 \city{Beijing}
 \country{China}
}

\author{Lixin Fan}
\orcid{0000-0002-8162-7096}
\email{lixinfan@webank.com}
\author{Haijun Yang}
\orcid{0009-0008-0491-7058}
\email{navyyang@webank.com}
\affiliation{
 \institution{WeBank}
 \city{Shenzhen}
 \country{China}
}

\author{Can Yang}
\orcid{0000-0002-4407-3055}
\email{macyang@ust.hk}
\affiliation{
 \institution{Hong Kong University of Science and Technology}
 \department{Mathematics}
 \city{Hong Kong}
 \country{China}
}

\author{Xiaowei Zhou}
\orcid{0000-0003-1926-5597}
\authornote{Corresponding author: Xiaowei Zhou.}
\email{xwzhou@zju.edu.cn}
\affiliation{
 \institution{State Key Laboratory of CAD\&CG, Zhejiang University}
 \city{Hangzhou}
 \country{China}
}

\begin{abstract}
    This paper addresses the challenging task of reconstructing the poses of multiple individuals engaged in close interactions, captured by multiple calibrated cameras. The difficulty arises from the noisy or false 2D keypoint detections due to inter-person occlusion, the heavy ambiguity in associating keypoints to individuals due to the close interactions, and the scarcity of training data as collecting and annotating motion data in crowded scenes is resource-intensive. 
    We introduce a novel system to address these challenges. Our system integrates a learning-based pose estimation component and its corresponding training and inference strategies. The pose estimation component takes multi-view 2D keypoint heatmaps as input and reconstructs the pose of each individual using a 3D conditional volumetric network.
    As the network doesn't need images as input, we can leverage known camera parameters from test scenes and a large quantity of existing motion capture data to synthesize massive training data that mimics the real data distribution in test scenes.
    Extensive experiments demonstrate that our approach significantly surpasses previous approaches in terms of pose accuracy and is generalizable across various camera setups and population sizes. 
    The code is available on our project page: \url{https://github.com/zju3dv/CloseMoCap}.
\end{abstract}

\newcommand{\shortname}{SAHMG }
\newcommand{\chatgpt}{ChatGPT }
\newcommand{\smplx}{SMPL-X }
\newcommand{\humanise}{HUMANISE }



\begin{teaserfigure}
    \includegraphics[width=\textwidth, trim=10 0 5 0,clip]{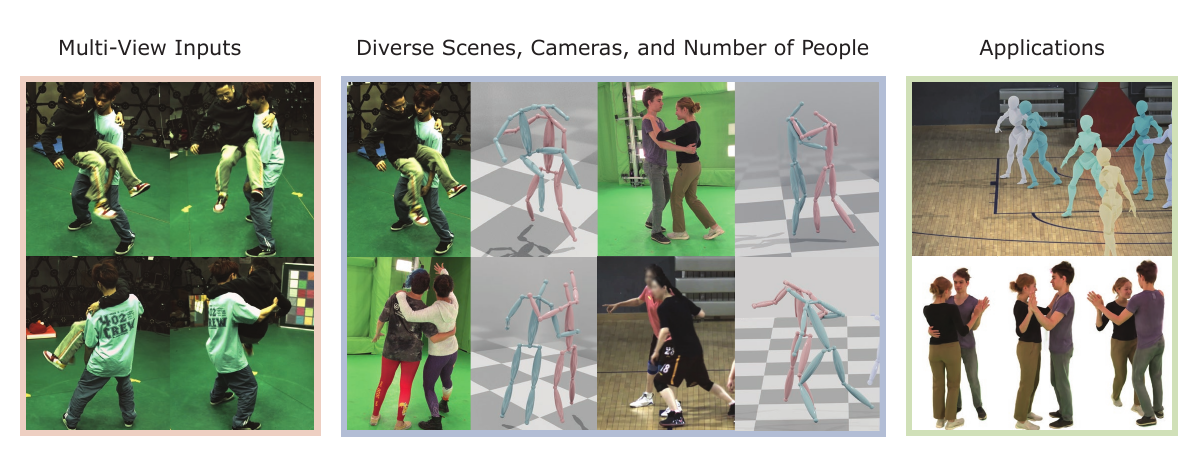}
    \caption{\textnormal{Our system is designed to recover the 3D poses of individuals engaging in close-range interactions, utilizing input from multiple calibrated cameras. We introduce a novel learning-based approach that effectively handles occlusions and interactions between individuals at close quarters. The standout feature of our system, which allows it to be trained without real data, enables the system to handle various scenes, camera configurations, and number of individuals. Our system facilitates a broad range of real applications, such as character animation (top-right) and free-viewpoint video synthesis (bottom-right).}}
    \label{fig:teaser}
  \end{teaserfigure}

\maketitle

\section{Introduction}
Markerless human motion capture is a vital enabling technology with broad applications spanning character animation, motion analysis, and 3D reconstruction of dynamic events. Compared to active or passive motion capture techniques, it has lower hardware requirements and fewer environmental constraints, capable of capturing human movement from a single or multiple calibrated cameras. This makes it more adaptable to a wide array of downstream needs. 

In this paper, our primary focus is the estimation of multi-person 3D poses from multiple calibrated cameras, particularly in scenarios involving close-range interactions. This is critical in many real scenarios, especially those that involve human-human interactions, such as multi-person dance performances and basketball games. These close-range interactions cause substantial challenges to pose estimation methods due to severe occlusions and heavy ambiguities when detecting and associating human body keypoints in multi-view images. 
Traditional methods~\cite{mvpose, 4d, quickpose} typically first estimate 2D poses from each viewpoint, then associate different 2D instances or keypoints across these views, and finally estimate 3D poses through triangulation. However, such association-based methods suffer from 2D estimation errors, especially in cases of close-range interaction. Additionally, the process of keypoint association becomes extremely challenging in crowded scenarios. This often results in either missed keypoints or low-precision pose estimates for these methods.

\begin{figure}[t]
\centering
\includegraphics[width=0.7\linewidth, trim=50 70 50 30, clip]{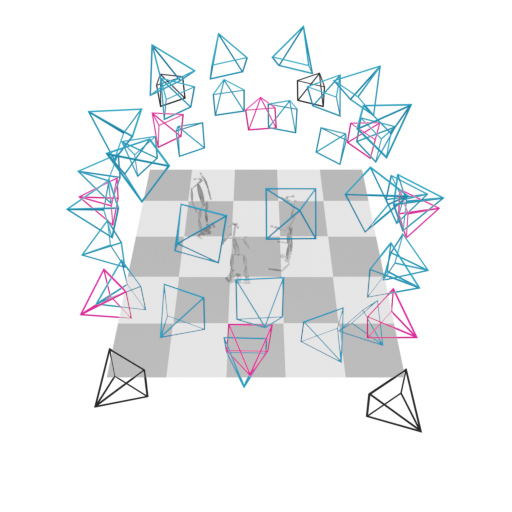}
\caption{\textbf{Diverse camera position and orientation in various datasets.} We show the camera positions and orientations of several common datasets: Panoptic~\cite{panoptic} (blue), CHI3D~\cite{chi3d} (black), and Hi4D~\cite{hi4d} (red). These datasets, collected by different studios, exhibit significant variations. In practical scenarios like a basketball court, scale differences become more apparent, posing challenges for network generalization across datasets.
}
\label{fig:illu_dataset}
\end{figure}

In contrast, learning-based approaches~\cite{voxelpose, fastervoxelpose, graph, direct} avoid performing 2D association. Instead of using keypoints as input directly, they construct feature volumes from multi-view 2D feature maps and then regress poses directly in 3D space in an end-to-end fashion. 
However, these learning-based methods heavily rely on paired 2D-3D ground-truth data for training. 
Existing datasets are typically acquired indoors using marker-based methods~\cite{h36m} or dense multi-view triangulation~\cite{panoptic} for ground-truth annotation. These datasets often lack diversity in terms of performers, actions, camera configurations, and background scenes. They also struggle to capture and annotate complex interactions due to severe occlusions. 
As a result, learning-based methods trained on these datasets are hard to generalize to different scenarios involving close interactions (as shown in Fig.~\ref{fig:illu_dataset}). The change in camera placement can significantly impact the outcomes, thereby limiting the generalization capability of learning-based methods in real-world applications.

\begin{figure}[t]
\centering
\includegraphics[width=0.4\linewidth, trim=200 150 200 0, clip]{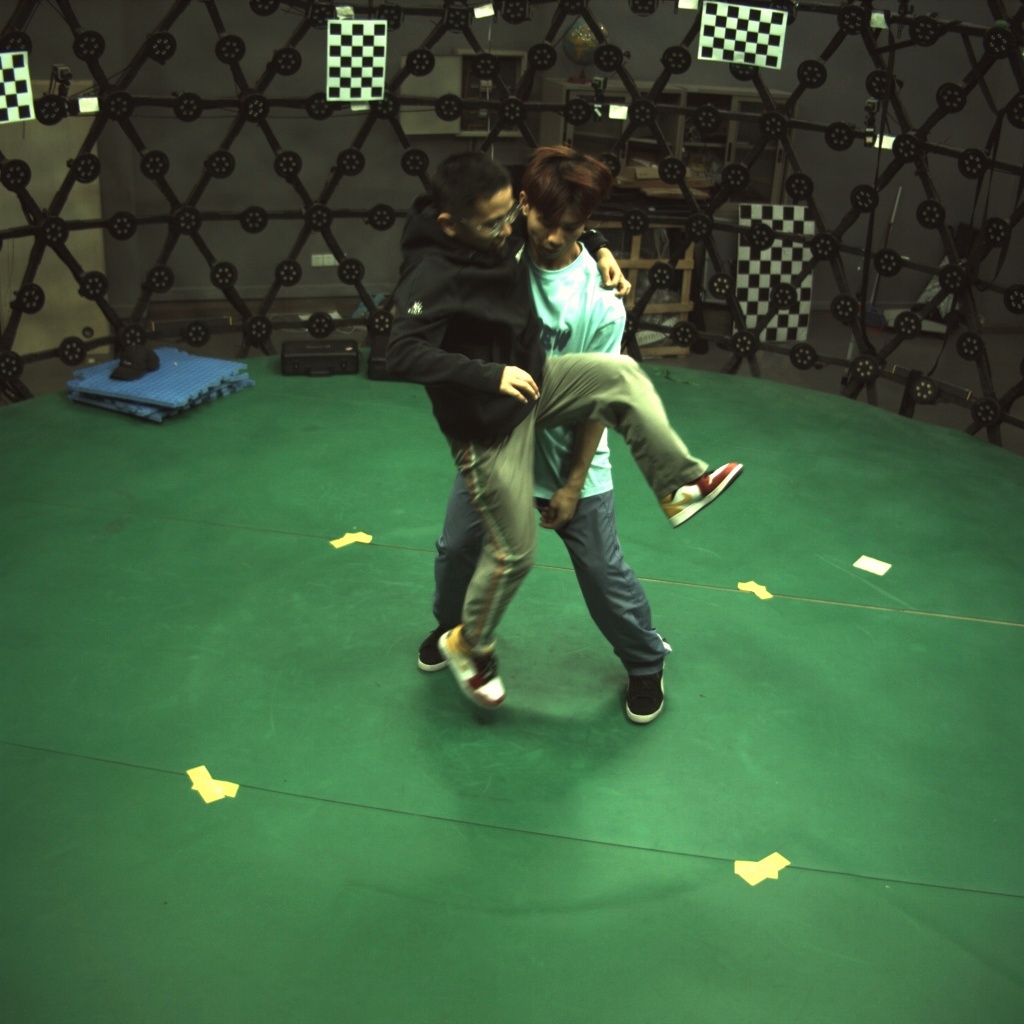}
\includegraphics[width=0.4\linewidth, trim=200 150 200 0, clip]{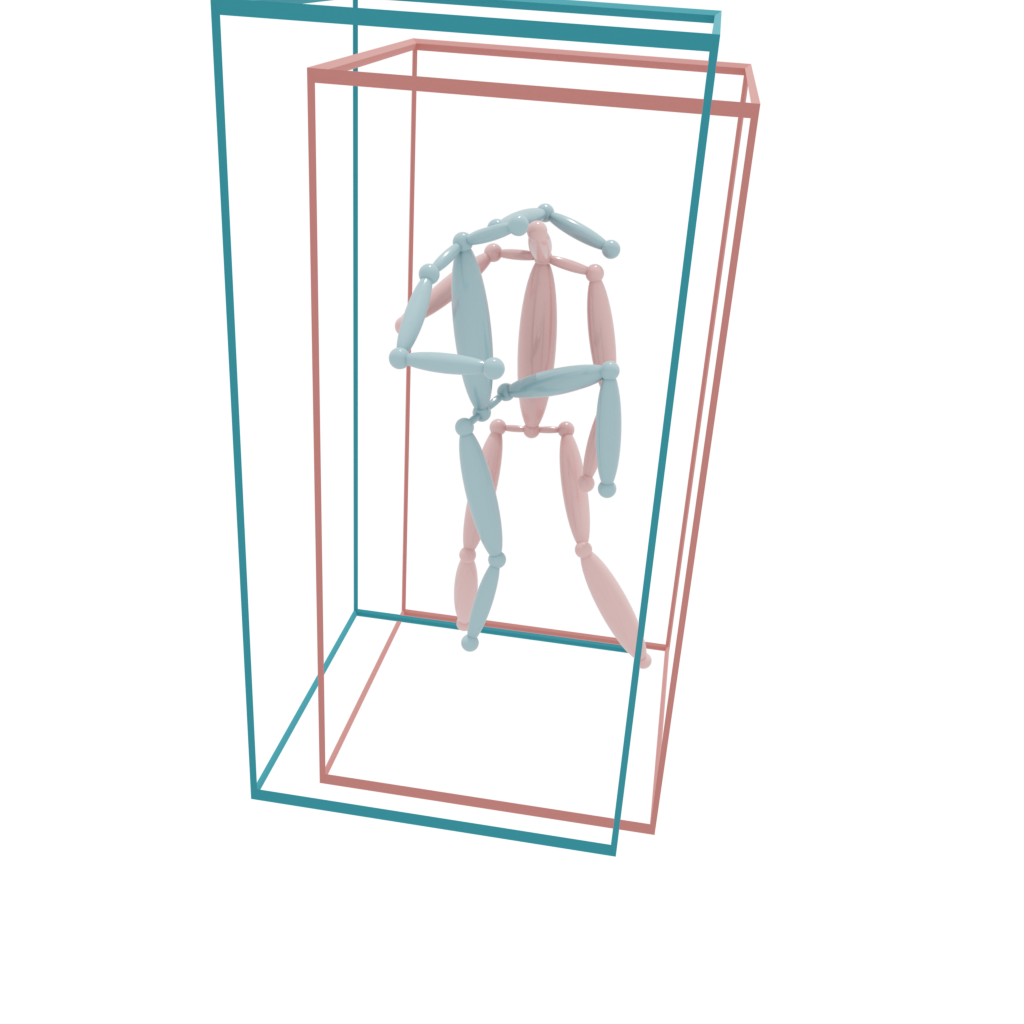}
\caption{\textbf{Challenges in pose estimation with close proximity.} This image highlights that when two individuals are in close proximity, it becomes difficult to obtain accurate 2D pose estimates due to heavy inter-person occlusion and keypoint association ambiguity. Moreover, in learning-based methods that directly regress 3D poses from feature volumes, the similarity in constructed volumes due to their spatial closeness complicates keypoint distinction for regression networks.}
\label{fig:ill_bbox3d}
\end{figure}

To overcome these challenges, we propose a novel system that is designed for close interaction scenarios and can be trained with only synthetic data.
The proposed method initially estimates keypoint heatmaps from multi-view images. Subsequently, it reconstructs the 3D centers for each individual. These centers are used to construct feature volumes, which are then passed through the 3D volumetric network. Finally, the network outputs the 3D pose of each individual.
Previous approaches often directly regress the 3D pose of each individual using the keypoint feature volume derived from 2D keypoint heatmaps. However, these methods struggle to represent close interactions among multiple individuals. When two individuals are in close proximity (as shown in Fig.~\ref{fig:ill_bbox3d}), their pelvis-centered feature volumes are highly similar, which may create ambiguity in the network output. To address this issue, we incorporate the estimated 3D centers for all individuals as an additional input to the network. This information is used to build the anchor-guided feature volumes, serving as a conditional signal.

Typically, image-based pose regression networks demand substantial quantities of multi-view image-3D pose pairs for training, which are often challenging to obtain. Fortunately, our method relies solely on 2D keypoints heatmaps as input. With known 3D poses and camera parameters, we can synthesize multi-view 2D heatmaps to train our network. For common multi-view tasks, we leverage existing camera parameters and a wealth of MoCap data to generate these 2D heatmaps. To enhance realism, we apply data augmentation to closely mimic heatmaps from real close interaction scenarios.
The vast MoCap dataset provides a rich set of motion data, improving network robustness. Besides, 2D heatmaps can be easily obtained during inference with off-the-shelf 2D pose estimators, making our method user-friendly and practical.

We conduct experiments on the latest close interaction datasets, which demonstrate that our method significantly outperforms previous approaches in terms of accuracy and robustness. We also validate our method across various scenarios with different scene scales, number of people, and camera configurations (see Fig.~\ref{fig:teaser}), affirming its robustness and applicability.

Our contributions are summarized as follows:
\begin{itemize}
    \item We propose a novel system to solve the problem of multi-person markerless motion capture in close interaction scenarios.
    \item We tackle the challenge of training data scarcity by generating synthetic samples using known camera parameters and extensive MoCap data.
    \item Through comprehensive experiments, we demonstrate our method's superior performance and applicability across various challenging scenarios.
\end{itemize}

\section{Related Works}

\paragraph{Multi-view multi-person 3D pose estimation}
Given calibrated multi-view images, existing approaches can be divided into two main categories: association-based methods and learning-based methods. Association-based methods usually take associated 2D poses as input and lift them into 3D poses. While learning-based methods usually associate and regress 3D poses from 2D keypoint heatmaps or detections with deep neural networks.

Association-based methods usually adopt a two-stage approach. In the first stage, 2D human poses are independently estimated for each view using pre-trained models~\cite{openpose, hrnet}. In the second stage, the estimated 2D poses of the same individual in different views are associated and lifted to 3D coordinates via triangulation~\cite{mvpose, dynamicmatching, quickpose}, plane sweep stereo~\cite{planesweep}, or the pictorial model~\cite{3dps,re3dps}. 
Existing methods mainly differ in the second stage.
\citeN{mvpose} grouped the 2D instances from geometric and appearance cues. 
\citeN{dynamicmatching} proposed a novel end-to-end training scheme including a dynamic matching algorithm for association.
\citeN{4d} proposed a 4D association method that first built a 4D graph from sequential 2D poses and applied a bundle Kruskal’s algorithm to search and assemble limbs based on the 4D graph. 
\citeN{planesweep} associated multi-view 2D poses based on plane sweep stereo and then applied 1D convolution neural networks (CNNs) to regress keypoints depth for 3D pose estimation.
\citeN{quickpose} considered all plausible skeletons via a tree-structure graph and reformulated the association problem as mode seeking.
However, it is hard for these methods to deal with missing or erroneous 2D keypoints.

Learning-based methods typically take 2D keypoints heatmaps or feature maps as input and convert them into 3D features. These 3D features are then used to regress multiple 3D poses directly using deep neural networks.
As a pioneer, VoxelPose~\cite{voxelpose} extended the learnable single-person pose estimator~\cite{lrtri} to multi-person scenarios. It first built 3D score volumes from multi-view 2D heatmaps. Then it used 3D CNNs to localize each individual from a coarse volume and estimate their 3D poses from fine volumes.
The follow-up work primarily enhanced VoxelPose in terms of speed and performance. 
Faster-VoxelPose~\cite{fastervoxelpose} realized real-time pose estimation by using BEV representation for localization and replacing computationally expensive 3D CNNs with efficient tri-plane 2D CNNs.
Meanwhile, \citeN{graph} applied graph convolution networks (GCNs) for human localization and pose regression, achieving significant improvements in terms of performance and computation complexity.
Taking a different approach, \citeN{direct} proposed a multi-view Transformer to predict multi-person 3D joint positions directly from multi-view images without the need for human localization.
Although these methods reduce the reliance on 2D pose estimation, they do not consider close human interactions.

\paragraph{Single-view multi-person 3D pose estimation}
Compared to multi-view inputs, single-view pose estimation presents a more accessible option, leading many methods to focus on directly estimating the poses of multiple people from a single image.
These methods can be primarily categorized into two classes: top-down and bottom-up. Top-down methods first perform human detection and then estimate 3D poses for each detected individual. 
\citeN{ROOTNET} first estimate the root depth of the human body and then the root-relative 3D pose. Subsequent works~\cite{pandanet,hmor,hdnet,das} improve estimation accuracy by considering factors like occlusion, depth relations, and joint distribution. 
Bottom-up methods~\cite{xnect,smap} first predict the 3D locations of all human keypoints and then associate them with each individual. 
Instead of 3D skeleton representation, more recent works~\cite{putting,mpik,glamr,slahmr,psvt,remips} have focused on recovering multiple SMPL~\cite{SMPL} or GHUM~\cite{ghum} meshes from monocular images or videos. 
Despite the remarkable progress in monocular 3D human pose estimation, these techniques still suffer from depth ambiguity and occlusion, making it difficult for them to obtain high-precision estimation. 
\begin{figure*}[t]
    \centering
    \includegraphics[width=0.99\textwidth, trim=0 0 20 0]{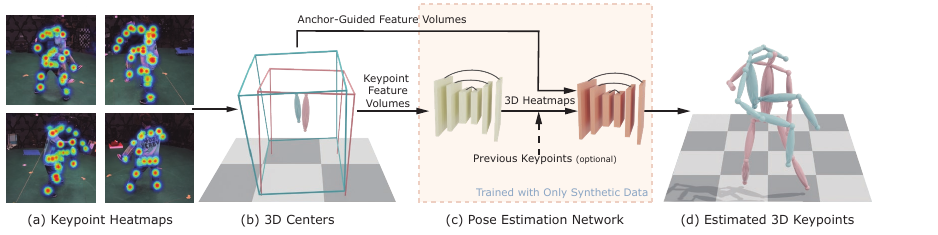}
    \caption{\textbf{Illustration of our method.} For a multi-view scene, we first estimate (a) the 2D keypoint heatmaps of all people from input images. We then recover (b) the 3D centers of all people from these heatmaps. Following this, we construct keypoint feature volumes and anchor-guided feature volumes, which are subsequently fed through (c) the pose estimation network. The proposed network initially predicts the 3D heatmaps from the keypoint feature volumes and then utilizes these 3D heatmaps along with the anchor-guided feature volumes to generate (d) the 3D keypoints for each person. If the 3D keypoints from the previous time step are available, they can be used to filter the 3D heatmaps. The entire network training does not require real image-3D keypoints pairs; instead, can be accomplished with only synthetic data. }
    \label{fig:pipeline}
\end{figure*}

\paragraph{3D human pose datasets}
Currently there exist many datasets providing RGB images and 3D pose annotations in both single-person and multi-person scenarios.
Given the ease of data collection, there are numerous single-person 2D~\cite{coco,eft,spin} and 3D datasets~\cite{humaneva,h36m,3DPW,humman,humbi,mono-3dhp2017,human4d,mhad,aifit,totalcapture,SCP}, featuring diverse actors, actions, and modalities. 
However, training and validation data are relatively scarce in multi-person scenarios.
Commonly used datasets such as Shelf and Campus~\cite{3dps} contain a limited number of frames, scenarios, and actions, making them less appropriate for comprehensive training and evaluation.
Panoptic~\cite{panoptic}, currently the largest real-world multi-person dataset, mainly focuses on social motion, ignoring more challenging cases, such as sports and close interactions. 

In recent years, several datasets containing close human interactions have been proposed. 
CHI3D~\cite{chi3d} and ExPI~\cite{expi}, for instance, captured numerous two-person interaction actions with multi-view cameras. However, they lack subject diversity as they only include a few pairs of actors. The most recent interaction dataset Hi4D~\cite{hi4d} registers neural implicit avatars to raw scans, enabling 4D tracking of both geometry and pose. However, all these datasets are restricted to indoor scenes.

Instead of capturing real data, some methods have attempted to apply synthetic data to overcome the issue of data deficiency.
Some methods~\cite{virtualpose,dsed} synthesize 2D heatmaps while most~\cite{surreal,muco,agora,HSPACE,BEDLAM} synthesize RGB images.
For instance, Surreal~\cite{surreal} utilizes MoCap poses~\cite{cmumocap} and textures~\cite{CAESAR} to synthesize a single-person dataset.
MuCO~\cite{muco} and AGORA~\cite{agora} use synthetic methods to generate static multi-person datasets.
To reduce sim-to-real gap, HSPACE~\cite{HSPACE} and BEDLAM~\cite{BEDLAM} synthesized large-scale photo-realistic dynamic video datasets with scan animation or physical simulation, but they only showed results on general scenes without close interactions.

Besides synthesizing data, some methods have applied view augmentation to improve generalizability. \citeN{3dfrom2d} transformed 2D pose into a 3D pose and random projected it into a 2D pose in a new viewpoint. The artificially projected 2D pose was then evaluated by a discriminator using an adversarial approach. \citeN{geoselfsup} extended \citeN{3dfrom2d} by lifting the projected 2D pose into a 3D pose, projecting it back to 2D with inverse transformation, and proposing a consistency loss for both 2D and 3D space. \citeN{canonpose} explored view augmentation on multi-view data.
However, these methods are not directly adaptable to multi-person scenarios.

\section{Technical approach}

Our system takes as input the multi-view images of a scene with known camera parameters and outputs the 3D poses of all individuals in the scene, particularly for cases involving close interactions. To achieve this, we begin by extracting 2D keypoint heatmaps and estimating the 3D center of each individual (Section \ref{sec-center}). Subsequently, we construct feature volumes for each person (Section \ref{sec-volume}) and feed them through our proposed network (Section \ref{sec-network}), which is trained with only synthetic data (Section \ref{sec-dataset}). The overview of our method is illustrated in Fig.~\ref{fig:pipeline}.

\subsection{Center Estimation and Tracking}\label{sec-center}

The center of each individual provides a rough estimate of their position within the scene. We use the pelvis point as the center of the body. From each image, 2D human keypoint heatmaps are extracted to identify potential 2D positions for the center points. Our goal is to select 2D candidates from different views and triangulate them to derive a valid 3D point. A valid 3D point should exhibit a minimal reprojection error when projected onto the selected views.

\paragraph{Triangulation from 2D candidates.} Once we have these 2D candidates, they are arranged in descending order based on their scores. Beginning by selecting the two candidates with the highest scores (heatmap responses) from different views, we proceed to triangulate them to create a 3D point. We then project this point to all views and check the reprojection error. If the error is below the specified threshold, we consider this 3D point as valid. Otherwise, we proceed to select the next 2D candidate with the highest score. This iterative process, which involves the selection, triangulation, and subsequent evaluation of candidate positions, continues until all the 2D candidate positions have been evaluated.

\paragraph{Tracking from previous frame.} For sequential input, we can streamline the procedure. We simply project the estimated centers of all individuals from the previous frame onto the current frame, retaining only those 2D points that meet the threshold condition. Subsequently, selected 2D points are triangulated to derive the center coordinates for the current frame. 
These reconstructed centers are utilized to construct feature volumes for each individual. Our method is also applicable to the neck point since both the pelvis and neck points can be robustly detected and tracked. These two points serve as anchor points for subsequent stages.

\subsection{Constructing Feature Volume}\label{sec-volume}

Previous methods usually discretize the 3D space into a fixed-size volume. They build a feature volume from 2D heatmaps or image features and employ a network to estimate the keypoint probabilities. In our approach, we strive for the network to be independent of actual image features. Hence, we opt to construct the feature volume using 2D heatmaps.

\begin{figure}[t]
\centering
\includegraphics[width=0.6\linewidth]{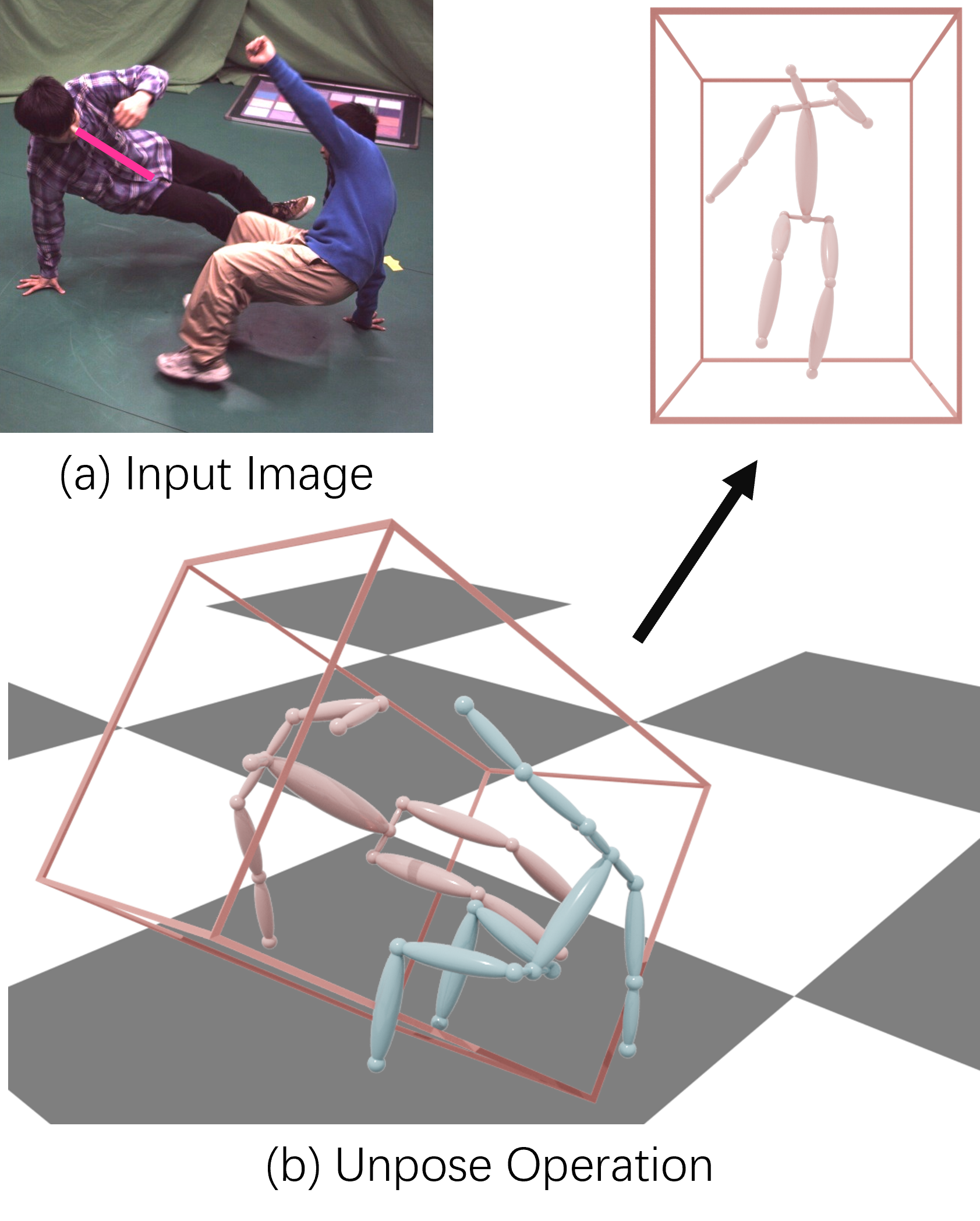}
\caption{\textbf{Coordinate transformation in 3D pose estimation.} We apply "unpose" operation to the estimated torso part (marked by the pink line in (a)), which is transformed into a standard space thus reducing the influence of global rotation.}
\label{fig:unpose}
\end{figure}

\paragraph{Unposing transformation.}
Before constructing the feature volume, we first "unpose" extreme poses encountered in real-world scenarios, such as lying down or rolling over, to a standard space (as illustrated in Fig.~\ref{fig:unpose}). 
The approach usually has minimal impact on common interaction datasets like CHI3D~\cite{chi3d} and Hi4D~\cite{hi4d}, as most of the poses in these datasets involve standing. However, it is significantly beneficial when dealing with more complex actions.
We use the estimated pelvis as the volume's center and the vector from the pelvis to the neck (as described in Sec.~\ref{sec-center}) as the reference vector. 
When constructing the volume, we rotate and translate the world coordinate system so that the transformed coordinate system is centered at the pelvis, with its z-axis aligned with the reference vector.
The keypoint position in the standard coordinate system is then converted back to the original world coordinates via the inverse rotation and translation.

\paragraph{Keypoint feature volume.} Given the unposed volume of each individual, we divide it into a discrete grid of size $W\times H\times D$. For each point $\mathbf{x}\in\mathbb{R}^3$ within this volume, we project it onto the $v$-th view using full-perspective projection and then bilinearly sample the heatmaps at its projected location. Subsequently, we combine the heatmap response vectors from different views into a global vector through averaging. The entire process can be formulated as:
\begin{equation}
    \mathbf{F}_{x} =\frac{1}{V}\sum_v\mathbf{h}_v\left(\Pi_v(\mathbf{x};\mathbf{K}_v,\mathbf{R}_v,\mathbf{t}_v)\right),
\end{equation}
where $\mathbf{K}_v$, $\mathbf{R}_v$, and $\mathbf{t}_v$ are cameras intrinsics and extrinsics of $v$-th viewpoint and $\Pi_v$ is the perspective projection function. Here, $\mathbf{h}_v$ represents the mapping function that samples the response of each joint from the 2D heatmap.

\paragraph{Anchor-guided feature volumes.} Simply building feature volumes around centers will introduce significant ambiguity when people closely interact with each other. This is because the feature volumes are nearly the same for each individual, as illustrated in Fig. \ref{fig:ill_bbox3d}. We address this ambiguity by employing the anchor points (pelvis and neck) since they provide positional information for the upper and lower body respectively. We utilize the anchor points of the $i$-th person, denoted as $\mathbf{c}^i\in\mathbb{R}^{2\times3}$, to extract the relevant features for that individual. Additionally, we use the anchor points of the other individuals, denoted as $\mathbf{c}^i_o = \{\mathbf{c}^k\in\mathbb{R}^{2\times3} | k = 1, \ldots, N, k \neq i\}$, to suppress the responses of keypoints belonging to others.

We model the response of grid points to the anchor points using a Gaussian function. The positive response volume $\mathbf{Z}^i \in\mathbb{R}^{2\times W\times H\times D}$ is calculated as follows:
\begin{equation}
    \mathbf{Z}^i = \exp{\left(-\frac{1}{2\sigma^2}\lvert\lvert\mathbf{x} - \mathbf{c}^i\rvert\rvert^2_2\right)},\label{torso-F}
\end{equation}
where $\sigma$ represents the Gaussian radius. To account for the negative response, we calculate $\mathbf{Z}^k\in\mathbb{R}^{2\times W\times H\times D}$ for each element of $\mathbf{c}^i_o$ using \equref{torso-F}. This is followed by an element-wise maximum operation to obtain a fused field $\mathbf{Z}^i_o\in\mathbb{R}^{2\times W\times H\times D}$ computed as:
\begin{equation}
    \mathbf{Z}^i_o=\operatorname{max}_k\mathbf{Z}^k,
\end{equation}
where $k$ denotes the index of other individuals. We call $\mathbf{Z}^i$ and $\mathbf{Z}^i_o$ \textit{anchor-guided feature volumes}.

The keypoint feature volume and the anchor-guided feature volumes serve as inputs to the pose estimation network. Both can be synthesized using 3D human poses, which means that we can train our network without the need for actual images.

\begin{figure}[t]
\centering
\includegraphics[width=1\linewidth]{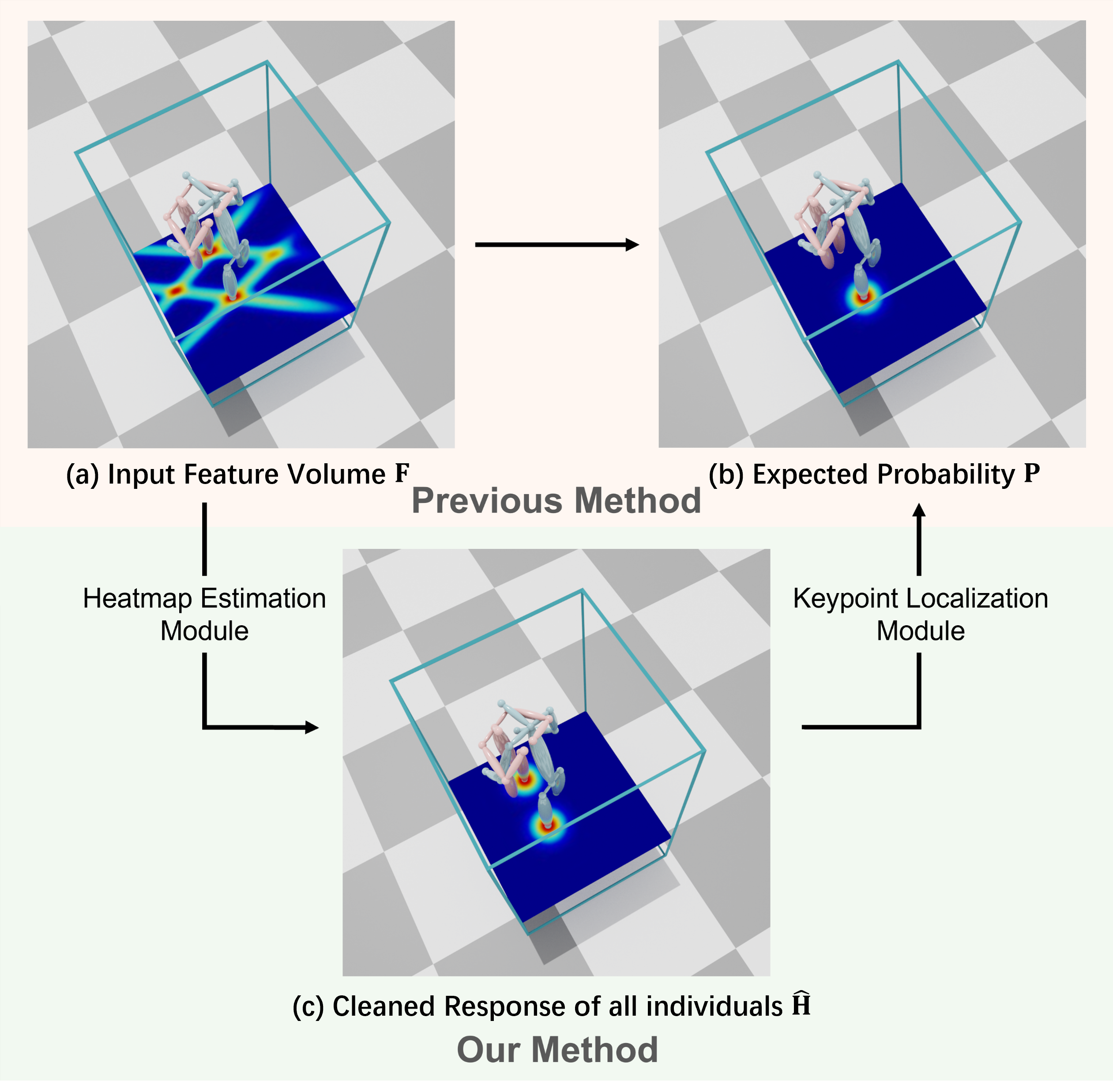}
\caption{\textbf{Two-stage design.} This image highlights the main difference between our approach and the previous methods in the field. Given the feature volume obtained through multiple viewpoints (a), the previous methods directly estimate the keypoint probability volume (b) of a target person. In contrast, we propose a two-stage method. The first Heatmap Estimation Module focuses on identifying and filtering out the noise present in the input feature volume and outputs a cleaned response volume of all individuals (c), while the second Keypoint Localization Module leverages the cleaned response volume and the conditional inputs to acquire the desired keypoint probability volume for each individual. This two-stage design allows the network to gain a better understanding of the scene.}
\label{fig:method_net}
\end{figure}

\subsection{Pose Estimation Network}\label{sec-network}
Previous approaches~\cite{voxelpose,fastervoxelpose} typically feed a keypoint feature volume $\mathbf{F}^i\in\mathbb{R}^{J\times W\times H\times D}$ of each person into the network to estimate the keypoint probability volume $\mathbf{P}^i\in\mathbb{R}^{J\times W\times H\times D}$. However, these methods struggle to accurately recover 3D poses in close interaction scenarios. This challenge often arises from the noise (multiple high responses for the same keypoint) and ambiguity (similar values for different people) present in the feature volume $\mathbf{F}^i$, as illustrated in Fig.~\ref{fig:method_net}. 

\paragraph{Two-stage network} To address these issues, we propose a 3D \textit{Heatmap Estimation Module} (HEM) and a \textit{Keypoint Localization Module} (KLM) to enhance the network's understanding of the scene. In the first stage, we take $\mathbf{F}^i$ as input and output a 3D heatmap volume $\mathbf{\hat{H}}^i\in\mathbb{R}^{J\times W\times H\times D}$ for all appeared keypoints:

\begin{equation}
    \mathbf{\Hat{H}}^i = \operatorname{CNN_{HEM}}(\mathbf{F}^i).
\end{equation}
This stage filters out erroneous responses and provides a well-regularized feature space for the subsequent stage. In the second stage, we concatenate $\mathbf{\hat{H}}^i$ with anchor-guided feature volumes and feed them into 3D CNNs to robustly regress the keypoint probability volume $\mathbf{P}^i$ for each person. The inclusion of the anchor-guided feature volumes helps the network learn to suppress keypoint responses from other individuals and focus on the target person.

The entire process can be formulated as:
\begin{equation}
    \mathbf{P}^i = \operatorname{CNN_{KLM}}\left(\mathbf{\hat{H}}^i,\mathbf{Z}^i,-\mathbf{Z}^i_o\right).
\end{equation}

The final 3D coordinates can be calculated by taking the expectation of this volume as follows:
\begin{equation}
    \mathbf{\hat{y}}^i_j = \sum_{l=1}^{W}\sum_{m=1}^{H}\sum_{n=1}^{D} \mathbf{P}^i_j(\mathbf{x})\cdot\mathbf{x}\label{eq2}.
\end{equation}

\paragraph{Temporal filtering.} For video input, we can further enhance the precision of the estimation by leveraging the estimated keypoints $\{\hat{\mathbf{y}}^i_{j,t-1}|j=0,\ldots,J\}$ from the previous frame. Given the 3D heatmap volume estimated by HEM, we eliminate the incorrect responses within it by assuming that the movement distance of the $j$-th keypoint between two consecutive frames is less than a threshold $r$. This step can be formally described as:
\begin{equation}
\hat{\mathbf{H}}^i_j(\mathbf{x}) = 
\begin{cases} 
\hat{\mathbf{H}}^i_j(\mathbf{x}) & \text{if } ||\hat{\mathbf{y}}^i_{j,t-1} - \mathbf{x}||_2 < r \\
0 & \text{otherwise,}
\end{cases}
\end{equation}
where $\mathbf{x}$ is the grid point of 3D heatmap volume. We set $r=0.05\text{m}$ in our experiments. 

The proposed pose estimation network can be trained end-to-end with 3D keypoint supervision. The temporal filtering module serves as a preprocessing step during inference.

\begin{figure}[t]
\centering
\includegraphics[width=1\linewidth]{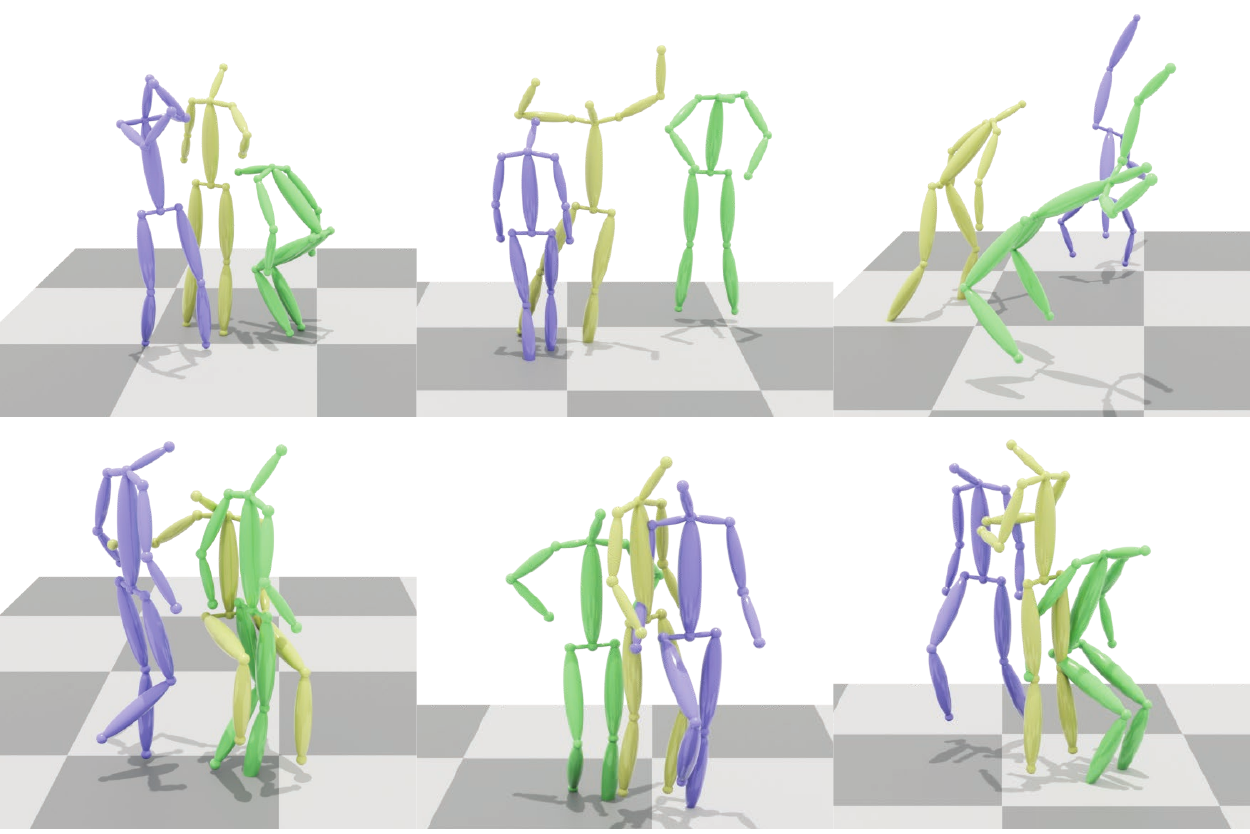}
\caption{\textbf{Synthesized multi-person 3D skeletons from the CMU MoCap~\cite{cmumocap} dataset.} The first row demonstrates some interactive actions sampled from the dataset, while the second row shows some synthesized close interactions.}
\label{fig:synthesis-3d}
\end{figure}

\subsection{Synthetic Training Data}\label{sec-dataset}

While attention has been given to collecting interaction data~\cite{chi3d,hi4d,expi}, real-world multi-person motion datasets with diverse subjects, actions, and settings remain limited. This limitation arises from the difficulty of data collection, resulting in a significant gap in diversity and richness compared to other tasks. To alleviate the data scarcity issue, we propose a training strategy that relies only synthetic data.

Given the camera parameters of a test scene, we randomly sample $N$ 3D poses from publicly available MoCap datasets and project them onto each camera plane to obtain the corresponding 2D heatmaps. This process results in a set of 2D-3D pose pairs. We perform data augmentation at both the 3D and 2D levels to improve data diversity. Specifically, for each sampled 3D pose, we randomly rotate and place it within the scene. To synthesize close interactions, we manually move the placed poses towards each other (as shown in Fig.~\ref{fig:synthesis-3d}). At the 2D level, we simulate keypoint occlusion using two strategies: random viewpoint dropout and random 2D keypoint dropout. 
To simulate the uncertainty of actual 2D backbone inference (\eg, missing keypoints, false detections, imperfect Gaussian distribution, and noise), we perform augmentation on the generated heatmaps by randomly perturbing the values, positions, and shapes of Gaussians.

We obtain the 3D motion data from the CMU MoCap~\cite{cmumocap} dataset, which contains motion clips of 543.49 minutes. To reduce data redundancy, we select 671,790 poses based on the measure of movements. The selected poses cover a wide range of human actions. The impact of data synthesis strategies is analyzed by experiments in Section \ref{sec:eval_synthetic}.

\section{Implementation Details}

\paragraph{Keypoint definition.} To be consistent with previous methods~\cite{voxelpose}, we preserve the same definition of keypoints as used in earlier works, \ie, the 15 keypoints defined in the CMU Panoptic dataset~\cite{panoptic}. Other datasets used in our study are converted to follow this definition for evaluation purposes, by using the given SMPL~\cite{SMPL} parameters and the SMPL model.

\paragraph{2D keypoint heatmap estimator.} We train the HRNet~\cite{hrnet} on the COCO~\cite{coco} dataset, requiring the network to output the keypoint responses of all individuals within a bounding box (consistent with the supervision of bottom-up methods). Off-the-shelf bottom-up 2D pose estimators (\eg, OpenPose~\cite{openpose}) can also be applied in our approach.

\paragraph{Pose estimation network.} Similar to previous works~\cite{voxelpose}, we utilize a voxel-to-voxel 3D convolutional network as the basic module. For each individual, we construct a volume of 2m$\times$2m$\times$2m, which is divided into a $W\times H\times D$ grid where $W=H=D=32$. 

\paragraph{Training.} 
We apply the Focal Loss~\cite{focalloss} for heatmap estimation in the first stage, 
and apply the L1 Loss to the regressed keypoint coordinates $\hat{\mathbf{y}}^i_j$ in the second stage.
The final loss function for the $i$-th person is computed as:
\begin{equation}
    L^i = \lambda \text{FocalLoss}(\mathbf{\hat{H}}^i, \mathbf{H}^i)+\frac{1}{J}\sum_j\lvert\mathbf{\hat{y}}^i_j - \mathbf{y}^i_j\rvert,\label{finalloss}
\end{equation}
where $\mathbf{H}^i$ is the ground-truth 3D keypoint heatmap of the $i$-th person, and $\mathbf{y}^i_j$ denotes the $j$-th ground truth keypoint coordinate of the $i$-th person. We use the Adam~\cite{adam} optimizer with a learning rate of 0.0001 and apply exponential decay with a gamma value of 0.95. We train the model on synthetic data for 50 epochs with a batch size of 32. For hyper-parameters, we use $\sigma=0.05\text{m}$ in \equref{torso-F}, and $\lambda=1$ in \equref{finalloss} for all experiments.

\section{Experiments}

\newcommand{\cmr}{ss}
\begin{table*}[ht]
    \begin{center}
        \caption{\textbf{Evaluation on CHI3D~\cite{chi3d}}. We report 3D Percentage of Correct Keypoints (3DPCK) with a threshold of 50mm here, so higher is better. Our approach achieves state-of-the-art results compared to the previous methods, surpassing even learning-based methods trained on the dataset by a large margin. `\textdagger' indicates the methods that are trained on Panoptic~\cite{panoptic} with 4 camera views close to CHI3D. `*' indicates the methods that are trained with synthetic data generated by their official code. `**' indicates the methods that are trained on the `s02' and `s04' sequences of CHI3D.}\label{tab:chi3d_eval}
    \resizebox{\textwidth}{!}{
    \footnotesize
    \begin{tabular}{llccccccccccc}
        \toprule
        & Method & Grab & Handshake & Hit & HoldingHands & Hug & Kick & Posing & Push & All \\
        \midrule
            \multirow{8}{*}{\makecell[c]{Learning-based\\methods}} & VoxelPose\textdagger       &36.28 &38.20 &41.60 &40.21 &31.22 &41.04 &26.08 &41.02 &38.36  \\
                           & Faster-VP\textdagger &10.22 &9.99 &9.72 &7.61 &7.58 &7.85 &6.79 &10.63 &9.19  \\
                           & Graph\textdagger               &28.70 &28.89 &30.97 &26.14 &20.40 &32.62 &18.00 &30.31 &28.33  \\ \cmidrule(lr){2-11}
         & VoxelPose$^{*}$                 &82.78 &86.55 &84.93 &89.01 &64.14 &84.47 &72.99 &86.82 &82.40  \\
                           & Faster-VP$^{*}$                 &81.91 &85.57 &84.66 &91.99 &65.03 &84.52 &71.59 &85.37 &82.22  \\ \cmidrule(lr){2-11}
        & VoxelPose$^{**}$                &90.79 &92.61 &91.41 &92.57 &75.30 &90.92 &78.51 &91.70 &88.91  \\
                           & Faster-VP$^{**}$                &89.93 &92.40 &90.35 &96.05 &67.42 &89.07 &77.48 &90.70 &87.45  \\
                           & Graph$^{**}$                    &93.45 &\textbf{95.77} &93.68 &95.80 &75.84 &93.63 &76.83 &93.88 &90.92  \\
        \midrule
        \multirow{2}{*}{\makecell[c]{Association-based\\methods}} & MVPose             &74.42 &77.65 &75.48 &84.48 &57.86 &75.01 &69.33 &76.55 &74.16 \\
                           & 4DA                    &79.80 &75.17 &76.65 &83.75 &72.83 &76.88 &78.02 &77.79 &77.45  \\
        \midrule
        &\textbf{Ours}     &\textbf{95.68} &95.28 &\textbf{94.92} &\textbf{97.40} &\textbf{88.14} &\textbf{94.30} &\textbf{92.16} &\textbf{95.23} &\textbf{94.30}  \\
        \bottomrule
    \end{tabular}
    }
    \end{center}
\end{table*}

\subsection{Datasets}

We use public datasets AMASS~\cite{amass}, CHI3D~\cite{chi3d}, Hi4D~\cite{hi4d} and CMU Panoptic~\cite{panoptic} for training and evaluation.

AMASS~\cite{amass} is a large and diverse database of human motion that includes multiple marker-based MoCap datasets. It uses MoSh++ to obtain the SMPL~\cite{SMPL} parameters for each data point. We primarily use the largest CMU MoCap~\cite{cmumocap} part as our training data for all experiments. We use the SMPL model to convert the SMPL parameters into 3D keypoints. This dataset contains 96 subjects and 1983 motions. We remove frames with movements less than 0.05m and ultimately obtain 671,790 frames in total. 

CHI3D~\cite{chi3d}
is an indoor dataset that focuses on close human interactions. It contains 631 multi-view sequences and 728,664 3D skeletons. Each sequence captures 2 people performing various actions from 4 different views with a resolution of 900 $\times$ 900 at 50 fps. The dataset also provides the ground-truth (GT) SMPL-X~\cite{SMPL-X} parameters for one person (obtained from markers) and pseudo GT parameters for the other person (obtained from RGB images). We use `s02' and `s04' as the training sequences for the baseline methods and `s03' as the test sequence for all methods since the actors in `s02' and `s04' are the same. We sample all sequences at 10fps and convert the provided SMPL-X parameters into 3D keypoints defined by COCO.

Hi4D~\cite{hi4d} 
is the latest indoor close interaction dataset containing 100 sequences. Each sequence captures 2 actors from 8 different viewpoints with a resolution of 1280 $\times$ 940 at 50 fps. The dataset provides the GT SMPL~\cite{SMPL} parameters for each frame. We select `fight00', `fight12', `hug00', `hug09', `hug12', `dance10', `dance14', `dance28', `pose32' and `pose37' for evaluation. We use all frames within the selection and convert the SMPL parameters into COCO keypoints similar to CHI3D. 

CMU Panoptic~\cite{panoptic}
is the largest real-world benchmark for multi-view multi-person 3D pose estimation. It contains 65 sequences and 1.5 million 3D skeletons with 30+ HD cameras. We use this dataset to evaluate our method on the setting of general multi-person poses. We follow \citeN{voxelpose} testing our method on `160906\_pizza1', `160422\_haggling1', `160906\_ian5', and `160906\_band4'.

\subsection{Baseline Methods}
We compare our method against the latest multi-view multi-person 3D pose estimation methods from two categories: learning-based methods and association-based methods. The learning-based baselines include the voxel-based method VoxelPose~\cite{voxelpose} and its fast version Faster-VP~\cite{fastervoxelpose}, as well as a graph-based method Graph~\cite{graph}. The association-based baselines include the top-down method MVPose~\cite{mvpose} and the bottom-up method 4DA~\cite{4d}.

\subsection{Metrics}
We utilize two metrics to evaluate the estimated 3D poses: 3D Percentage of Correct Keypoints (3DPCK) [\%] and Mean Per Joint Position Error (MPJPE) [mm]. 3DPCK determines whether a keypoint is correctly estimated by measuring the Euclidean distance between the ground-truth (GT) position and the estimated joint position. If this distance is within a threshold, the estimation is considered as correct. Specifically, we match each estimated pose with the closest GT pose. If multiple estimates match with the same GT, only the one with the highest proposal score is viewed as a True Positive (TP), while the others are viewed as False Positives (FP). 
MPJPE calculates the Euclidean distance between the GT keypoint positions and the matched estimated keypoint positions. 

\subsection{Evaluation on Close-Interaction Datasets}

The quantitative evaluation of our method and the baseline methods on the CHI3D~\cite{chi3d} dataset is given in Tab.~\ref{tab:chi3d_eval}.

\begin{figure*}[htbp]
    \centering
    \newlength{\qualcol}
    \setlength{\qualcol}{0.14\textwidth}
    \begin{tabular}{@{\ }p{\qualcol}@{\ }p{\qualcol}@{\ }p{\qualcol}@{\ }p{\qualcol}@{\ }p{\qualcol}@{\ }p{\qualcol}}
        \multicolumn{6}{c}{\includegraphics[width=6\qualcol]{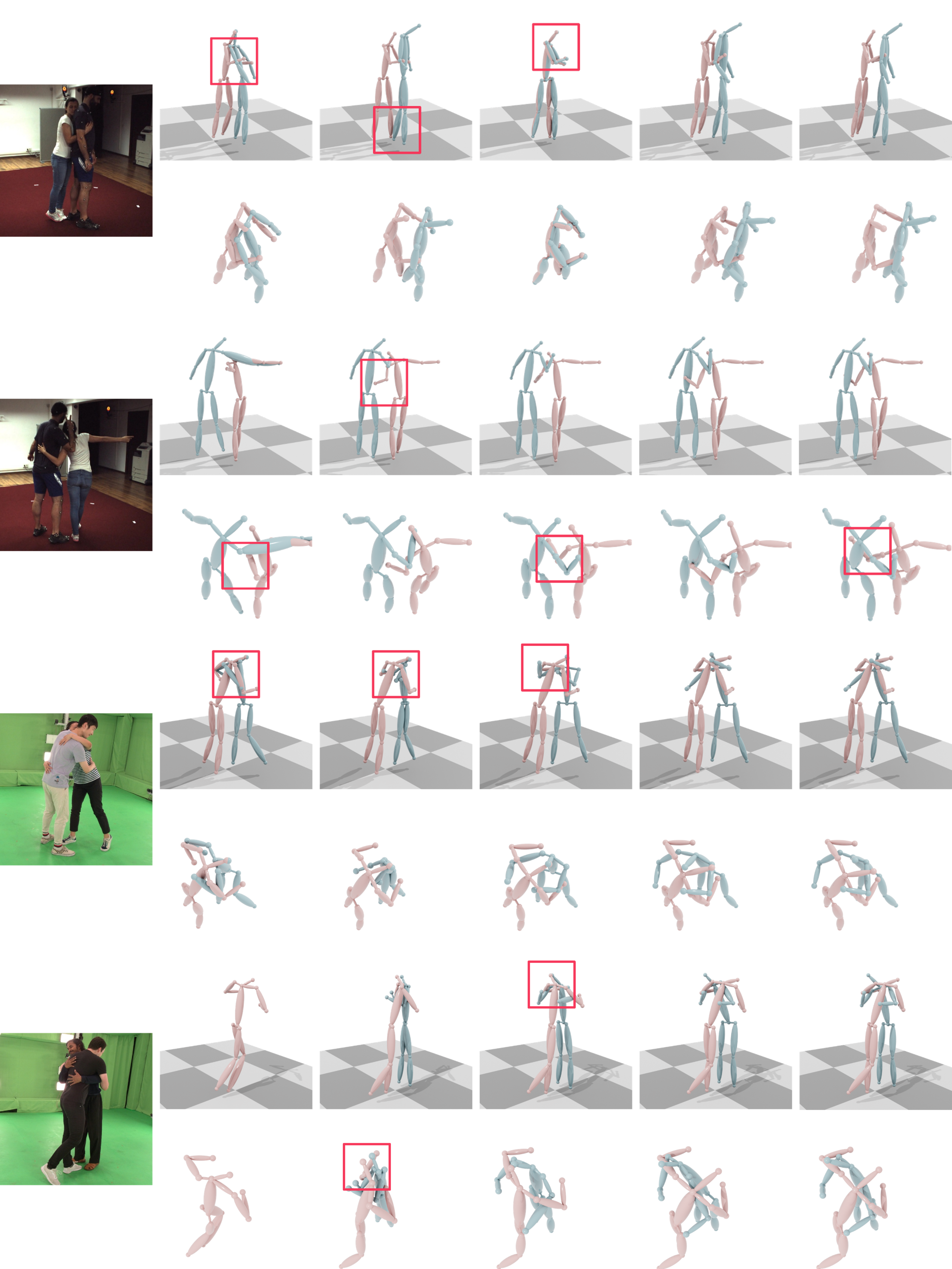}} \\
        \makecell[c]{\small } &
        \makecell[c]{\small VoxelPose} &
        \makecell[c]{\small Faster-VP} &
        \makecell[c]{\small Graph} &
        \makecell[c]{\small Ours} &
        \makecell[c]{\small GT}
    \end{tabular}
    \caption{\textbf{Qualitative comparison with learning-based methods on CHI3D~\cite{chi3d} and Hi4D~\cite{hi4d}}. Our method exhibits superior accuracy in reconstructing 3D poses without using any real paired training data compared to other learning-based methods, particularly in challenging scenarios. The red box indicates the erroneous keypoints estimated by learning-based methods and even existed in the pseudo ground truth in CHI3D.
    }
    \label{fig:chi3d_hi4d_learing_vis}
\end{figure*}

\paragraph{Comparison with learning-based methods} %
Learning-based methods directly trained on the Panoptic~\cite{panoptic} dataset exhibit poor performance on CHI3D. The results suggest that these methods struggle to generalize across different camera layouts. Therefore, we train VoxelPose~\cite{voxelpose} and Faster-VP~\cite{fastervoxelpose} with synthetic heatmaps generated by their official code given the cameras of CHI3D, obtaining improved results. The best performance of learning-based methods is obtained by training using the real image-pose pairs from CHI3D. This implies that learning-based methods perform better with a sufficient amount of training data from the same scene.
Compared to these methods, our method significantly outperforms them, demonstrating the effectiveness of our approach. Some qualitative comparisons are shown in Fig. \ref{fig:chi3d_hi4d_learing_vis}.

\begin{figure}[t]
    \includegraphics[width=\linewidth]{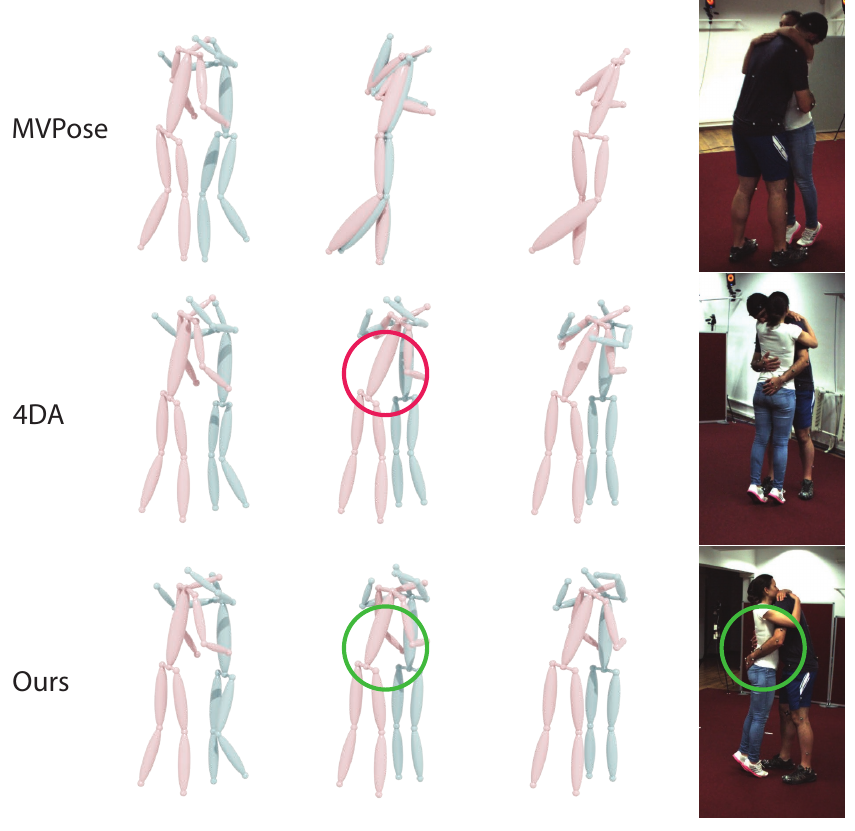} 
    \centering
        
    \caption{\textbf{Qualitative comparison with association-based methods.} This figure compares the efficacy of our method and two pose estimation methods at varying interaction distances. We render skeletons using the viewpoint of the image in the first row. The traditional top-down method MVPose performs well at long distances but fails at close range. The bottom-up method 4DA excels at close range, though it still fails to reconstruct some keypoints as shown in the red circle. In contrast, our method accurately reconstructs poses in this complex scenario, outperforming the other two methods.}
    \label{fig:chi3d_trakcing_vis}
\end{figure}

\paragraph{Comparison with association-based methods}
MVPose~\cite{mvpose} relies on body-level matching and reconstruction, so it often struggles with close interactions. In contrast, 4DA~\cite{4d} initially reconstructs joints and then associates them in 3D, thus producing more reasonable skeletons, but its accuracy is still limited due to missing 2D detections. Compared with them, our method excels in reconstructing accurate poses in complex scenarios, as demonstrated in Fig. \ref{fig:chi3d_trakcing_vis}.

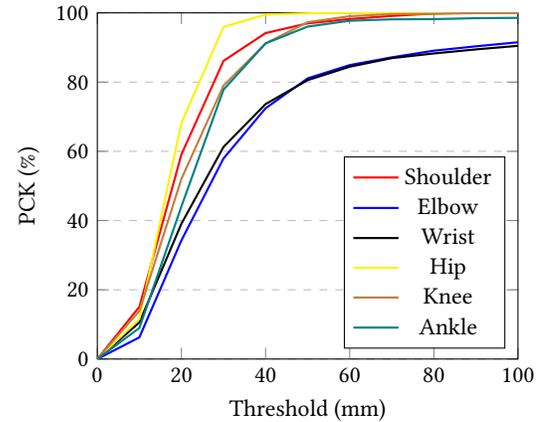
\begin{figure}[t]
    \begin{minipage}[htbp]{0.4\textwidth}
    \centering
    \begin{tikzpicture}
        \begin{axis}[
            name=plot1,
            xlabel={Threshold (mm)},
            ylabel={PCK (\%)},
            xmin=0,xmax=100,
            ymin=0,ymax=100,
            xtick={0,20,40,60,80,100},
            ytick={0,20,40,60,80,100},
            legend pos=south east,
            ymajorgrids=true,
            grid style=dashed,
            width=\textwidth,
            cycle list name=color list,
            ]
        \addplot
            coordinates {
                (0,0)
                (10,15.09)
                (20,59.11)
                (30,86.12)
                (40,94.16)
                (50,96.96)
                (60,98.22)
                (70,99.11)
                (80,99.81)
                (90,100.00)
                (100,100.00)
            };
        \addlegendentry{Shoulder}
        \addplot
            coordinates {
                (0,0)
                (10,6.26)
                (20,34.30)
                (30,57.94)
                (40,72.43)
                (50,81.07)
                (60,84.91)
                (70,87.15)
                (80,89.07)
                (90,90.33)
                (100,91.50)
            };
        \addlegendentry{Elbow}
        \addplot
            coordinates {
                (0,0)
                (10,10.56)
                (20,39.11)
                (30,61.26)
                (40,73.64)
                (50,80.56)
                (60,84.44)
                (70,86.92)
                (80,88.27)
                (90,89.44)
                (100,90.47)
            };
        \addlegendentry{Wrist}
        \addplot
            coordinates {
                (0,0)
                (10,11.59)
                (20,68.27)
                (30,95.93)
                (40,99.49)
                (50,99.86)
                (60,100.0)
                (70,100)
                (80,100)
                (90,100)
                (100,100)
            };
        \addlegendentry{Hip}
        \addplot
            coordinates {
                (0,0)
                (10,13.88)
                (20,52.06)
                (30,78.97)
                (40,91.17)
                (50,97.29)
                (60,99.02)
                (70,99.63)
                (80,99.72)
                (90,99.91)
                (100,99.95)
            };
        \addlegendentry{Knee}
        \addplot
            coordinates {
                (0,0)
                (10,8.97)
                (20,44.11)
                (30,77.80)
                (40,91.17)
                (50,96.03)
                (60,97.71)
                (70,98.13)
                (80,98.18)
                (90,98.46)
                (100,98.55)
            };
        \addlegendentry{Ankle}
        \end{axis}
    \end{tikzpicture}
    \end{minipage}
    \caption{\textbf{Keypoint-level evaluation on the `Hug' of CHI3D~\cite{chi3d}.} We report 3DPCKs with a threshold from 0 to 100mm on `Hug'. We can find that the 3DPCKs of `Elbow' and `Wrist' are larger than others. This is because the upper body experiences closer interaction than the lower body.}\label{fig:chi3d_auc}
\end{figure}

\paragraph{Keypoint-level evaluations}
We evaluate our method on different keypoints separately to further investigate the accuracy of interacting body parts. Specifically, we evaluate 3DPCKs of keypoints that are more involved in human-human interactions (\eg, Shoulder, Elbow, Wrist, Hip, Knee, and Ankle) on a typical scenario `Hug', as shown in Fig. \ref{fig:chi3d_auc}. We find that the overall accuracy of lower-body keypoints (\eg, Hip, Knee, and Ankle) is better than upper-body keypoints (\eg, Shoulder, Elbow, and Wrist). This is because the hugging motion mostly involves upper-body parts. 

\begin{figure}[t]
    \begin{minipage}[htbp]{0.4\textwidth}
    \centering
    \begin{tikzpicture}
        \begin{axis}[
            name=plot1,
            xlabel={Threshold (mm)},
            ylabel={PCK (\%)},
            xmin=0,xmax=100,
            ymin=0,ymax=100,
            xtick={0,20,40,60,80,100},
            ytick={0,20,40,60,80,100},
            legend pos=south east,
            ymajorgrids=true,
            grid style=dashed,
            width=\textwidth,
            legend style={font=\footnotesize},
            cycle list name=color list,
            ]
        \addplot
            coordinates {
                (0,0)
                (10,2.64)
                (20,18.13)
                (30,41.55)
                (40,58.65)
                (50,69.08)
                (60,74.77)
                (70,77.87)
                (80,80.30)
                (90,82.22)
                (100,83.61)
            };
        \addlegendentry{VoxelPose$^\S$}
        \addplot
            coordinates {
                (0,0)
                (10,3.82)
                (20,20.03)
                (30,43.27)
                (40,61.16)
                (50,71.63)
                (60,77.02)
                (70,80.28)
                (80,82.57)
                (90,84.44)
                (100,85.75)
            };
        \addlegendentry{Faster-VP$^\S$}
        \addplot
            coordinates {
                (0,0)
                (10,6.04)
                (20,30.08)
                (30,55.67)
                (40,71.92)
                (50,79.86)
                (60,84.01)
                (70,86.59)
                (80,88.26)
                (90,89.39)
                (100,90.38)
            };
        \addlegendentry{Graph$^\S$}
        \addplot
            coordinates {
                (0,0)
                (10,11.22)
                (20,43.07)
                (30,64.81)
                (40,75.34)
                (50,81.46)
                (60,84.53)
                (70,85.95)
                (80,86.98)
                (90,87.77)
                (100,88.28)
            };
        \addlegendentry{VoxelPose$^*$}
        \addplot
            coordinates {
                (0,0)
                (10,11.93)
                (20,42.77)
                (30,67.73)
                (40,78.12)
                (50,83.56)
                (60,86.36)
                (70,87.83)
                (80,88.82)
                (90,89.50)
                (100,89.96)
            };
        \addlegendentry{Faster-VP$^*$}
        \addplot
            coordinates {
                (0,0)
                (10,8.99)
                (20,38.46)
                (30,61.33)
                (40,73.98)
                (50,81.35)
                (60,84.77)
                (70,87.35)
                (80,89.47)
                (90,90.66)
                (100,91.35)
            };
        \addlegendentry{MVPose}
        \addplot
            coordinates {
                (0,0)
                (10,9.73)
                (20,36.82)
                (30,59.44)
                (40,76.08)
                (50,87.83)
                (60,93.39)
                (70,96.16)
                (80,97.21)
                (90,97.78)
                (100,98.24)
            };
        \addlegendentry{4DA}
        \addplot
            coordinates {
                (0,0)
                (10,18.17)
                (20,63.07)
                (30,86.66)
                (40,95.02)
                (50,98.22)
                (60,99.19)
                (70,99.37)
                (80,99.48)
                (90,99.56)
                (100,99.64)
            };
        \addlegendentry{Ours}
        \end{axis}
    \end{tikzpicture}
    \end{minipage}
    \caption{\textbf{Evaluation on Hi4D~\cite{hi4d} with tight thresholds}. We report 3DPCKs with a tight threshold from 0 to 100mm. The results show that our method outperforms others by a large margin even in tight thresholds. The notations of methods follow those in Tab. \ref{tab:hi4d_eval}.}\label{fig:hi4d_auc}
\end{figure}
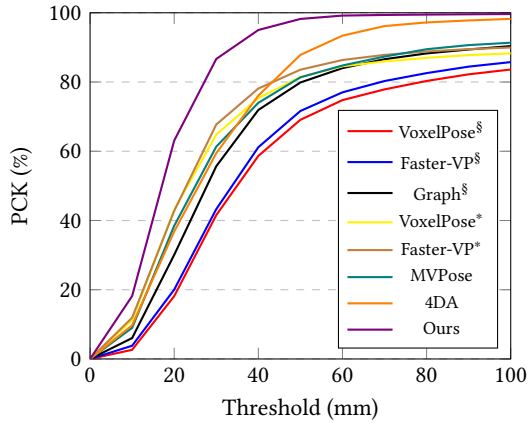

\paragraph{Generalization across datasets}
We further evaluate different methods on the Hi4D~\cite{hi4d} dataset without training on this dataset. The quantitative evaluation is given in Tab.~\ref{tab:hi4d_eval}. Learning-based methods trained on CHI3D exhibit similar performances, slightly worse than association-based methods. This suggests that with more cameras (8 cameras in Hi4D versus 4 cameras in CHI3D), association-based methods can outperform learning-based ones. Despite this, our method continues to surpass both learning-based and association-based methods. Additionally, we present the 3DPCK curves with thresholds ranging from 0-100mm for all methods. As shown in Fig. \ref{fig:hi4d_auc}, our method consistently outperforms others across all thresholds.

\begin{table}[ht]
    \caption{\textbf{Evaluation on Hi4D~\cite{hi4d}.} We report 3DPCKs with different thresholds (50mm, 100mm, and 200mm) and MPJPE.
    `\S' indicates the methods that are trained on CHI3D~\cite{chi3d} and tested on Hi4D using the 4 views close to the training ones. `*' indicates the methods that are trained with synthetic data generated from CHI3D using their official code.}\label{tab:hi4d_eval}
    \small
    \begin{tabular}{lcccc}
        \toprule
        Method & PCK@50$\uparrow$ & PCK@100$\uparrow$ & PCK@200$\uparrow$ & MPJPE$\downarrow$ \\
        \midrule 
        VoxelPose$^\S$      &69.08 &83.61 &89.41 &59.47  \\
        Faster-VP$^\S$      &71.63 &85.75 &93.21 &58.22  \\
        Graph$^\S$          &79.86 &90.38 &94.28 &45.63  \\
        \midrule
        VoxelPose$^*$       &83.56 &89.96 &92.28 &42.00  \\
        Faster-VP$^*$       &81.35 &91.35 &94.35 &43.07  \\
        \midrule
        4DA                  &87.83 &98.24 & 99.48 & 31.60  \\
        MVPose              &83.75 &95.87 &97.94 &37.53  \\
        \midrule
        \textbf{Ours}       &\textbf{98.29} &\textbf{99.55} &\textbf{99.68}  & \textbf{20.28}\\
        \midrule
        Ours(w/ $\mathbf{Y}_{t-1})$       & 98.22 & 99.64 & 99.90  & 19.40 \\
        \bottomrule
    \end{tabular}
\end{table}

\subsection{Evaluation on General Multi-Person Tracking Dataset}

To demonstrate the generality of our approach, we evaluate it on the Panoptic~\cite{panoptic} dataset which captures more regular social activities in a studio and serves as a common benchmark for previous research~\cite{voxelpose,fastervoxelpose,graph}. 
The results are given in Tab.~\ref{tab:panoptic}, where the numbers of the previous methods are adopted from their papers~\cite{voxelpose,fastervoxelpose,graph,direct} using their original data partition and evaluation protocols.
As this dataset contains very few examples of close interaction, previous methods also perform well on it. 
Our method achieves comparable performance even without the use of real image-3D pose pairs during training. Graph~\cite{graph} and MvP~\cite{direct} exhibit better results in the $\text{AP}_{25}$ metric. This improvement may be attributed to their use of multi-layered image features extracted by the 2D CNN. 
If we train our network using the keypoint heatmaps extracted from the real images, our AP$_{25}$ exhibits an improvement (Ours* in the table), where there is a marginal change in AP$_{50}$ and AP$_{100}$. This suggests that incorporating real heatmaps for training can enhance the network's precision.

\begin{table}[tbp]
\centering
\caption{\textbf{Evaluation on Panoptic~\cite{panoptic}.} We report Average Precision (AP) with thresholds of 25, 50, and 100mm, where the higher values indicate better performance. Ours is trained using synthetic data, while Ours* is trained using heatmaps generated from images.}\label{tab:panoptic}
\small
\begin{tabular}{clcccc}
\toprule
 & Method & $\text{AP}_{25}\uparrow$ & $\text{AP}_{50}\uparrow$ & $\text{AP}_{100}\uparrow$ \\
\midrule
\multirow{4}{*}{\makecell[c]{Heatmap-based\\methods}} & VoxelPose & 83.59 & 98.33 & \textbf{99.76} \\
& Faster-VP & 85.22 & 98.08 & 99.32 \\
& Ours & 86.16 & \textbf{98.70} & 99.49 \\
& Ours* & \textbf{90.10} & 98.57 & 99.34 \\ %
\midrule
\multirow{2}{*}{\makecell[c]{Image-feature-based\\methods}} & Graph & \textbf{94.00} & \textbf{98.93} & \textbf{99.76} \\
& MvP & 92.28 & 96.60 & 97.45 \\
\bottomrule
\end{tabular}
\end{table}

\subsection{Ablation Study}
 
\paragraph{Core components.}
We conduct ablation studies to evaluate the impact of our core components, \ie, \textit{3D heatmap supervision} and \textit{conditional inputs}. The experiments are evaluated on the CHI3D~\cite{chi3d} dataset. We use ground-truth centers here to mitigate the influence of center estimation. The results are shown in Tab.~\ref{tab-abla-network}. For \textit{3D heatmap supervision}, we train an ablated model without supervision of the 3D heatmap volume, \ie, the network is not explicitly required to output 3D heatmaps of all individuals. We find that the removal of heatmap supervision decreases the performance in all metrics, indicating the importance of heatmap supervision.
As for \textit{conditional inputs}, we train an ablated model where conditional inputs, \ie, the anchor-guided feature volumes, in the keypoint localization module are eliminated. The results show that the removal of conditional inputs leads to degraded performance, emphasizing the role of the conditional inputs in solving the ambiguity arising from the close interaction.

\begin{table}[ht]
\centering
\caption{\textbf{Ablation study on CHI3D~\cite{chi3d}.} We report the MPJPE, PCK@50 for all scenarios, and PCK@50 for the `Hug' scenario. This table highlights the significance of 3D heatmap supervision and conditional inputs and also shows the superior performance achieved with ground-truth 2D heatmaps.
}
\begin{small}
\begin{tabular}{lrrr}
\toprule
 & MPJPE$\downarrow$ & PCK$\uparrow$ & PCK@50(Hug)$\uparrow$ \\
\midrule
w/o 3D Heatmap Supervision & 24.53 & 91.58 & 85.05 \\
w/o Conditional Inputs & 22.02 & 95.56 & 90.42 \\
w GT 2D Heatmap & 8.77 & 99.92 & 99.46 \\
\midrule
Ours & 18.78 & 97.12 & 93.28 \\
\bottomrule
\end{tabular}
\end{small}
\label{tab-abla-network}
\end{table}

\paragraph{Heatmap used during inference.} 
We explore the performance difference between using GT heatmaps and estimated heatmaps during the inference phase in Tab.~\ref{tab-abla-network}. Experimental results show a noticeable improvement when using GT heatmaps during inference, indicating that the heatmap estimation of our pre-trained model still has room for improvement. Enhancements to the 2D heatmap estimation can effectively boost the overall performance of our approach.

\paragraph{Number of views.} 
We evaluate the impact of the number of views on the Hi4D~\cite{hi4d} dataset, as shown in Tab.~\ref{tab:comp-diffview}. As the number of viewpoints decreases, the error of the estimated center and the overall MPJPE increase.
Using the GT center estimation as an input reduces the error to a certain extent. This suggests that more accurate center estimation results in better overall performance.

\begin{table}[ht]
\centering
\caption{\textbf{Comparison of pose estimation performance with different numbers of views on Hi4D~\cite{hi4d}.} This table compares the performance of our proposed 3D pose estimation method with different numbers of views (4, 6, and 8). The performance metrics are MPJPE and MPJPE with ground-truth root (MPJPE w/ GT root). The lower values indicate better performance.}
\begin{tabular}{lccc}
\hline
 & Center Error$\downarrow$ & MPJPE$\downarrow$ & MPJPE w/ GT root$\downarrow$ \\ 
\hline
4 views & 29.78 & 28.85 & 24.30 \\
6 views & 23.01 & 21.50 & 16.07 \\
8 views & 20.16 & 19.22 & 13.79 \\
\hline
\end{tabular}
\label{tab:comp-diffview}
\end{table}

\subsection{Evaluation of Synthetic Training Strategies}
\label{sec:eval_synthetic}

We conduct experiments on the CHI3D~\cite{chi3d} dataset to evaluate the influence of our synthetic training strategies. The results are shown in Tab.~\ref{tab:ablation-data}. 

\paragraph{Quantity of synthetic data.}

To investigate the impact of synthetic data size on model performance, we train our model using only half or a quarter of the CMU MoCap~\cite{cmumocap} dataset. Additionally, we test the performance of using the combined CMU MoCap and the BMLMoVi~\cite{Movi} data from the AMASS~\cite{amass} dataset for training. The experiments show that using less training data leads to performance decline, but the overall performance is still close to that of using the full data. Adding the extra BMLMoVi data does not bring a significant performance improvement, which suggests that the CMU MoCap dataset provides a sufficient variety of motion data.

\paragraph{Different number of subjects in the scene.}

We sample two subjects from CHI3D during training. If we sample only one subject per frame, there is a significant performance drop. If we sample more subjects each time (\eg, five subjects), the performance is close to that of two subjects.

\paragraph{Data augmentation.} We further evaluate the effectiveness of 2D and 3D data augmentation. Without heatmap augmentation, the performance on the test set decreases significantly. If the center augmentation is not used, the performance degrades slightly.

\begin{table}[ht]
\centering
\caption{\textbf{Comparison between multiple synthetic training strategies on CHI3D~\cite{chi3d}.} This table demonstrates that the CMU MoCap\cite{cmumocap} dataset provides sufficient diversity for close-interaction pose estimation. Both multi-person sampling during synthetic data generation and heatmap augmentation are beneficial for our method.}\label{tab:ablation-data}
\begin{tabular}{lcc}
\toprule
 & MPJPE$\downarrow$ & PCK$\uparrow$ \\
\midrule
CMU MoCap + MoVi & 18.92 & \textbf{97.17} \\
1/2 CMU MoCap & 18.96 & 97.03 \\
1/4 CMU MoCap & 19.50 & 96.87 \\
\# of subjects = 1 & 23.97 & 93.73 \\
\# of subjects = 5 & 18.80 & 97.13 \\
w/o 2D heatmap augmentation & 51.99 & 76.81 \\
w/o center augmentation & 18.87 & 97.04 \\
\midrule
CMU MoCap + \# of subjects = 2 & \textbf{18.78} & 97.12 \\
\bottomrule
\end{tabular}
\end{table}

\subsection{Applicability to Large-Scale Scenes}

To verify the generalizability of our method to large-scale scenes, we evaluate it on a dataset from a basketball court. This dataset was collected using 28 calibrated RGB cameras that fully encircle the basketball court. The data represents a typical basketball game with 10 players.

\begin{figure*}[htbp]
    \centering
    \includegraphics[width=0.9\textwidth, trim=0 1250 0 200, clip]{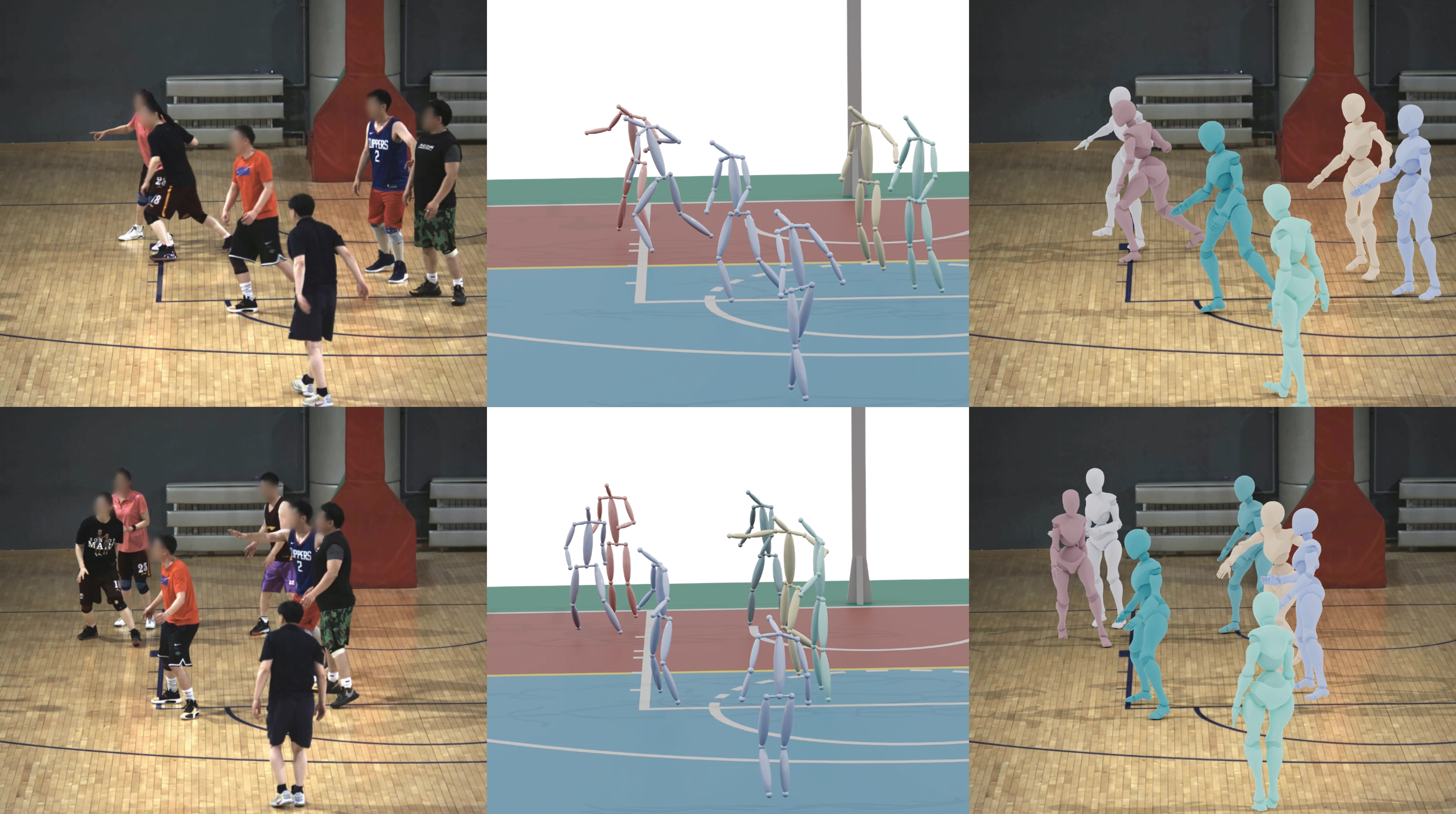} 
    \caption{This figure presents the results of our pose estimation and character animation on a large basketball court. Compared to indoor data, the basketball data has a larger scale and more people. Our method can perform pose estimation without the need for labeled training data on the basketball court. The results of pose estimation can be used for downstream applications such as motion analysis, character animation, and augmented reality.}
    \label{fig:res-basketball}
\end{figure*}

Our method can be easily applied to this scenario, using the known camera parameters to synthesize training data. To cope with the issue of 2D human estimation in large scenes during testing, we use an off-the-shelf object deterctor~\cite{yolov5} to obtain 2D human centers in the entire image, from which the 3D human centers are reconstructed. Then, we project the 3D bounding box of each person back to the image, followed by cropping the original image. The cropped images are then used to estimate the 2D heatmaps, which serve as input for the our pose estimation network. The input images and reconstruction results are shown in Fig.~\ref{fig:res-basketball}. For such a large-scale scene, our method is also able to handle close-range interactions between people effectively. 

\begin{figure*}[htbp]
    \centering
    \includegraphics[width=0.9\textwidth]{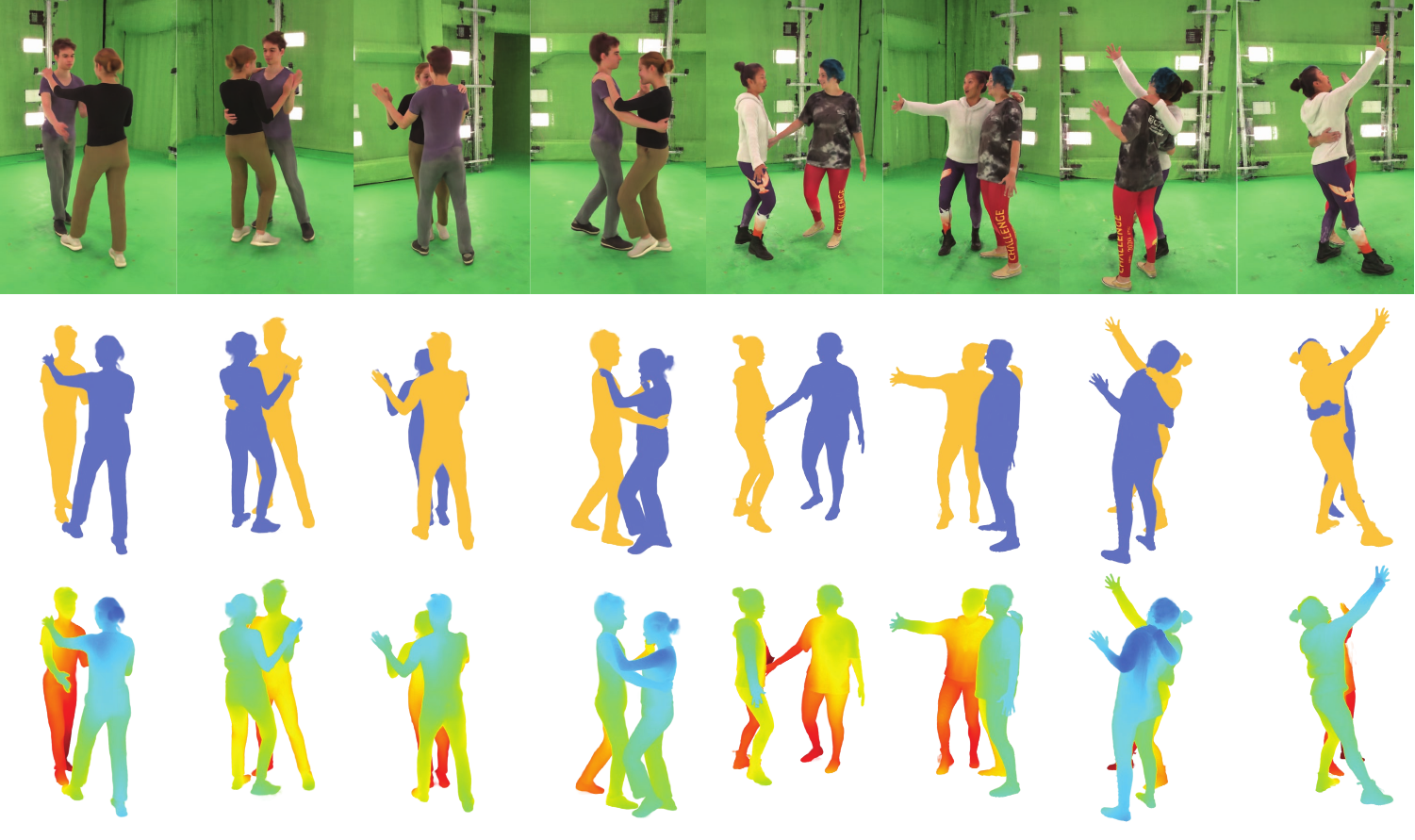} 
    \caption{This figure showcases the results of free-viewpoint rendering achieved by learning a neural radiance field~\cite{multinb} on the Hi4D~\cite{hi4d} dataset using eight viewpoints of RGB images and our estimated 3D poses. The experimental results demonstrate that our method can effectively assist downstream tasks, including novel view synthesis (top), instance segmentation (middle), and depth estimation (bottom).}
    \label{fig:res-mnb}
\end{figure*}

\section{Application in Novel View Synthesis}

Our method furnishes an advantageous starting point for many downstream tasks. Recently, novel view synthesis of dynamic humans from sparse views has been explored in \cite{neuralbody,neuralactor,humannerf,keypointnerf,multinb}. These methods adopted the SMPL \cite{SMPL} model or keypoints as the geometric prior to associate appearance information across different frames and learned a NeRF \cite{nerf} for rendering. We use the recent SMPL-based method \cite{multinb} for demonstration. The 3D skeletons recovered by our method are directly used to fit SMPL models. Subsequently, we employ multi-view images along with the SMPL parameters to train a dynamic radiance field following \cite{multinb}, thereby facilitating the rendering of novel views, instance masks and depth maps of the scene, as demonstrated in Fig. \ref{fig:res-mnb}. Contrasting with the data acquisition technique of the Hi4D dataset~\cite{hi4d}, our method requires merely eight RGB cameras, obviating the necessity for a high-cost scanning system. This simplifies the data acquisition process significantly while producing high-quality 3D reconstruction.

\section{Limitation and Future Work}

While our method has shown promising results, there are some limitations that open up avenues for further improvements. First, our method currently only takes 2D keypoint heatmaps as input to directly output 3D keypoint coordinates. This could be enhanced by incorporating more 2D features into the input, such as the Part Affinity Field (PAF) from OpenPose~\cite{openpose}. This additional feature provides limb link information, thus potentially improving pose estimation, particularly in complex multi-person scenarios.
Another limitation lies in the expressiveness of our output. Our method currently only outputs body keypoints, which may not fully capture the intricacy of human motion.
As such, future work could consider fitting human body models from the estimated 3D keypoints with surface or contact losses and incorporating additional hand and facial keypoints.
Lastly, our method does not involve motion prior during the training phase. We only utilize temporal input during the inference stage. Future work could incorporate temporal motion prior learning~\cite{humor} and spatial motion prior learning~\cite{sociallymotion, dyadicmotion,expi}. Such enhancement could potentially allow the model to gain more understanding of motion patterns and improve the overall accuracy of the pose estimation.

\section{Conclusion}

This paper introduces a novel system for multi-person 3D pose estimation from multi-view images, with a specific focus on scenarios involving close interactions. Our technical contribution lies in the conditional 3D volumetric network along with its corresponding training and inference strategies. We demonstrated through rigorous experiments that our method significantly outperforms previous approaches, exhibiting robustness and generalization across a variety of scenarios. We expect this work will facilitate more future research in related areas and real applications that require motion capture or dynamic reconstruction of closely interacting humans.

\begin{acks}   
    The authors would like to acknowledge support from NSFC (No. 62172364) and the Information Technology Center and State Key Lab of CAD\&CG, Zhejiang University.
\end{acks}

\bibliographystyle{ACM-Reference-Format}
\bibliography{bibliography}


\begin{thebibliography}{74}


\ifx \showCODEN    \undefined \def \showCODEN     #1{\unskip}     \fi
\ifx \showDOI      \undefined \def \showDOI       #1{#1}\fi
\ifx \showISBNx    \undefined \def \showISBNx     #1{\unskip}     \fi
\ifx \showISBNxiii \undefined \def \showISBNxiii  #1{\unskip}     \fi
\ifx \showISSN     \undefined \def \showISSN      #1{\unskip}     \fi
\ifx \showLCCN     \undefined \def \showLCCN      #1{\unskip}     \fi
\ifx \shownote     \undefined \def \shownote      #1{#1}          \fi
\ifx \showarticletitle \undefined \def \showarticletitle #1{#1}   \fi
\ifx \showURL      \undefined \def \showURL       {\relax}        \fi
\providecommand\bibfield[2]{#2}
\providecommand\bibinfo[2]{#2}
\providecommand\natexlab[1]{#1}
\providecommand\showeprint[2][]{arXiv:#2}

\bibitem[Adeli et~al\mbox{.}(2020)]%
        {sociallymotion}
\bibfield{author}{\bibinfo{person}{Vida Adeli}, \bibinfo{person}{Ehsan Adeli}, \bibinfo{person}{Ian Reid}, \bibinfo{person}{Juan~Carlos Niebles}, {and} \bibinfo{person}{Hamid Rezatofighi}.} \bibinfo{year}{2020}\natexlab{}.
\newblock \showarticletitle{Socially and contextually aware human motion and pose forecasting}.
\newblock \bibinfo{journal}{\emph{IEEE Robotics and Automation Letters}} \bibinfo{volume}{5}, \bibinfo{number}{4} (\bibinfo{year}{2020}), \bibinfo{pages}{6033--6040}.
\newblock


\bibitem[Bazavan et~al\mbox{.}(2021)]%
        {HSPACE}
\bibfield{author}{\bibinfo{person}{Eduard~Gabriel Bazavan}, \bibinfo{person}{Andrei Zanfir}, \bibinfo{person}{Mihai Zanfir}, \bibinfo{person}{William~T Freeman}, \bibinfo{person}{Rahul Sukthankar}, {and} \bibinfo{person}{Cristian Sminchisescu}.} \bibinfo{year}{2021}\natexlab{}.
\newblock \showarticletitle{HSPACE: Synthetic parametric humans animated in complex environments}.
\newblock \bibinfo{journal}{\emph{arXiv preprint arXiv:2112.12867}} (\bibinfo{year}{2021}).
\newblock


\bibitem[Belagiannis et~al\mbox{.}(2014)]%
        {3dps}
\bibfield{author}{\bibinfo{person}{Vasileios Belagiannis}, \bibinfo{person}{Sikandar Amin}, \bibinfo{person}{Mykhaylo Andriluka}, \bibinfo{person}{Bernt Schiele}, \bibinfo{person}{Nassir Navab}, {and} \bibinfo{person}{Slobodan Ilic}.} \bibinfo{year}{2014}\natexlab{}.
\newblock \showarticletitle{3D pictorial structures for multiple human pose estimation}. In \bibinfo{booktitle}{\emph{CVPR}}. \bibinfo{pages}{1669--1676}.
\newblock


\bibitem[Belagiannis et~al\mbox{.}(2015)]%
        {re3dps}
\bibfield{author}{\bibinfo{person}{Vasileios Belagiannis}, \bibinfo{person}{Sikandar Amin}, \bibinfo{person}{Mykhaylo Andriluka}, \bibinfo{person}{Bernt Schiele}, \bibinfo{person}{Nassir Navab}, {and} \bibinfo{person}{Slobodan Ilic}.} \bibinfo{year}{2015}\natexlab{}.
\newblock \showarticletitle{3d pictorial structures revisited: Multiple human pose estimation}.
\newblock \bibinfo{journal}{\emph{IEEE Transactions on Pattern Analysis and Machine Intelligence}} \bibinfo{volume}{38}, \bibinfo{number}{10} (\bibinfo{year}{2015}), \bibinfo{pages}{1929--1942}.
\newblock


\bibitem[Benzine et~al\mbox{.}(2020)]%
        {pandanet}
\bibfield{author}{\bibinfo{person}{Abdallah Benzine}, \bibinfo{person}{Florian Chabot}, \bibinfo{person}{Bertrand Luvison}, \bibinfo{person}{Quoc~Cuong Pham}, {and} \bibinfo{person}{Catherine Achard}.} \bibinfo{year}{2020}\natexlab{}.
\newblock \showarticletitle{Pandanet: Anchor-based single-shot multi-person 3d pose estimation}. In \bibinfo{booktitle}{\emph{CVPR}}. \bibinfo{pages}{6856--6865}.
\newblock


\bibitem[Black et~al\mbox{.}(2023)]%
        {BEDLAM}
\bibfield{author}{\bibinfo{person}{Michael~J. Black}, \bibinfo{person}{Priyanka Patel}, \bibinfo{person}{Joachim Tesch}, {and} \bibinfo{person}{Jinlong Yang}.} \bibinfo{year}{2023}\natexlab{}.
\newblock \showarticletitle{{BEDLAM}: A Synthetic Dataset of Bodies Exhibiting Detailed Lifelike Animated Motion}. In \bibinfo{booktitle}{\emph{CVPR}}. \bibinfo{pages}{8726--8737}.
\newblock


\bibitem[Cai et~al\mbox{.}(2022)]%
        {humman}
\bibfield{author}{\bibinfo{person}{Zhongang Cai}, \bibinfo{person}{Daxuan Ren}, \bibinfo{person}{Ailing Zeng}, \bibinfo{person}{Zhengyu Lin}, \bibinfo{person}{Tao Yu}, \bibinfo{person}{Wenjia Wang}, \bibinfo{person}{Xiangyu Fan}, \bibinfo{person}{Yang Gao}, \bibinfo{person}{Yifan Yu}, \bibinfo{person}{Liang Pan}, \bibinfo{person}{Fangzhou Hong}, \bibinfo{person}{Mingyuan Zhang}, \bibinfo{person}{Chen~Change Loy}, \bibinfo{person}{Lei Yang}, {and} \bibinfo{person}{Ziwei Liu}.} \bibinfo{year}{2022}\natexlab{}.
\newblock \showarticletitle{{HuMMan}: Multi-modal 4d human dataset for versatile sensing and modeling}. In \bibinfo{booktitle}{\emph{ECCV}}. Springer, \bibinfo{pages}{557--577}.
\newblock


\bibitem[Cao et~al\mbox{.}(2017)]%
        {openpose}
\bibfield{author}{\bibinfo{person}{Zhe Cao}, \bibinfo{person}{Tomas Simon}, \bibinfo{person}{Shih-En Wei}, {and} \bibinfo{person}{Yaser Sheikh}.} \bibinfo{year}{2017}\natexlab{}.
\newblock \showarticletitle{Realtime multi-person 2d pose estimation using part affinity fields}. In \bibinfo{booktitle}{\emph{CVPR}}. \bibinfo{pages}{7291--7299}.
\newblock


\bibitem[Cha et~al\mbox{.}(2022)]%
        {mpik}
\bibfield{author}{\bibinfo{person}{Junuk Cha}, \bibinfo{person}{Muhammad Saqlain}, \bibinfo{person}{GeonU Kim}, \bibinfo{person}{Mingyu Shin}, {and} \bibinfo{person}{Seungryul Baek}.} \bibinfo{year}{2022}\natexlab{}.
\newblock \showarticletitle{Multi-Person 3D Pose and Shape Estimation via Inverse Kinematics and Refinement}. In \bibinfo{booktitle}{\emph{ECCV}}. Springer, \bibinfo{pages}{660--677}.
\newblock


\bibitem[Chatzitofis et~al\mbox{.}(2020)]%
        {human4d}
\bibfield{author}{\bibinfo{person}{Anargyros Chatzitofis}, \bibinfo{person}{Leonidas Saroglou}, \bibinfo{person}{Prodromos Boutis}, \bibinfo{person}{Petros Drakoulis}, \bibinfo{person}{Nikolaos Zioulis}, \bibinfo{person}{Shishir Subramanyam}, \bibinfo{person}{Bart Kevelham}, \bibinfo{person}{Caecilia Charbonnier}, \bibinfo{person}{Pablo Cesar}, \bibinfo{person}{Dimitrios Zarpalas}, {et~al\mbox{.}}} \bibinfo{year}{2020}\natexlab{}.
\newblock \showarticletitle{HUMAN4D: A Human-Centric Multimodal Dataset for Motions and Immersive Media}.
\newblock \bibinfo{journal}{\emph{IEEE Access}}  \bibinfo{volume}{8} (\bibinfo{year}{2020}), \bibinfo{pages}{176241--176262}.
\newblock


\bibitem[Chen et~al\mbox{.}(2019)]%
        {geoselfsup}
\bibfield{author}{\bibinfo{person}{Ching-Hang Chen}, \bibinfo{person}{Ambrish Tyagi}, \bibinfo{person}{Amit Agrawal}, \bibinfo{person}{Dylan Drover}, \bibinfo{person}{Rohith Mv}, \bibinfo{person}{Stefan Stojanov}, {and} \bibinfo{person}{James~M Rehg}.} \bibinfo{year}{2019}\natexlab{}.
\newblock \showarticletitle{Unsupervised 3d pose estimation with geometric self-supervision}. In \bibinfo{booktitle}{\emph{CVPR}}. \bibinfo{pages}{5714--5724}.
\newblock


\bibitem[{CMU Graphics Lab}(2000)]%
        {cmumocap}
\bibfield{author}{\bibinfo{person}{{CMU Graphics Lab}}.} \bibinfo{year}{2000}\natexlab{}.
\newblock \bibinfo{title}{CMU Graphics Lab Motion Capture Database}.
\newblock
\newblock
\newblock
\shownote{\url{http://mocap.cs.cmu.edu/}}.


\bibitem[Dong et~al\mbox{.}(2019)]%
        {mvpose}
\bibfield{author}{\bibinfo{person}{Junting Dong}, \bibinfo{person}{Wen Jiang}, \bibinfo{person}{Qixing Huang}, \bibinfo{person}{Hujun Bao}, {and} \bibinfo{person}{Xiaowei Zhou}.} \bibinfo{year}{2019}\natexlab{}.
\newblock \showarticletitle{Fast and robust multi-person 3d pose estimation from multiple views}. In \bibinfo{booktitle}{\emph{CVPR}}. \bibinfo{pages}{7792--7801}.
\newblock


\bibitem[Drover et~al\mbox{.}(2018)]%
        {3dfrom2d}
\bibfield{author}{\bibinfo{person}{Dylan Drover}, \bibinfo{person}{Rohith MV}, \bibinfo{person}{Ching-Hang Chen}, \bibinfo{person}{Amit Agrawal}, \bibinfo{person}{Ambrish Tyagi}, {and} \bibinfo{person}{Cong Phuoc~Huynh}.} \bibinfo{year}{2018}\natexlab{}.
\newblock \showarticletitle{Can 3d pose be learned from 2d projections alone?}. In \bibinfo{booktitle}{\emph{ECCVW}}. \bibinfo{pages}{78--94}.
\newblock


\bibitem[Fieraru et~al\mbox{.}(2020)]%
        {chi3d}
\bibfield{author}{\bibinfo{person}{Mihai Fieraru}, \bibinfo{person}{Mihai Zanfir}, \bibinfo{person}{Elisabeta Oneata}, \bibinfo{person}{Alin-Ionut Popa}, \bibinfo{person}{Vlad Olaru}, {and} \bibinfo{person}{Cristian Sminchisescu}.} \bibinfo{year}{2020}\natexlab{}.
\newblock \showarticletitle{Three-dimensional reconstruction of human interactions}. In \bibinfo{booktitle}{\emph{CVPR}}. \bibinfo{pages}{7214--7223}.
\newblock


\bibitem[Fieraru et~al\mbox{.}(2021a)]%
        {SCP}
\bibfield{author}{\bibinfo{person}{Mihai Fieraru}, \bibinfo{person}{Mihai Zanfir}, \bibinfo{person}{Elisabeta Oneata}, \bibinfo{person}{Alin-Ionut Popa}, \bibinfo{person}{Vlad Olaru}, {and} \bibinfo{person}{Cristian Sminchisescu}.} \bibinfo{year}{2021}\natexlab{a}.
\newblock \showarticletitle{Learning complex 3d human self-contact}. In \bibinfo{booktitle}{\emph{AAAI}}. \bibinfo{pages}{1343--1351}.
\newblock


\bibitem[Fieraru et~al\mbox{.}(2021b)]%
        {aifit}
\bibfield{author}{\bibinfo{person}{Mihai Fieraru}, \bibinfo{person}{Mihai Zanfir}, \bibinfo{person}{Silviu-Cristian Pirlea}, \bibinfo{person}{Vlad Olaru}, {and} \bibinfo{person}{Cristian Sminchisescu}.} \bibinfo{year}{2021}\natexlab{b}.
\newblock \showarticletitle{AIFit: Automatic 3D Human-Interpretable Feedback Models for Fitness Training}. In \bibinfo{booktitle}{\emph{CVPR}}. \bibinfo{pages}{9919--9928}.
\newblock


\bibitem[Fieraru et~al\mbox{.}(2021c)]%
        {remips}
\bibfield{author}{\bibinfo{person}{Mihai Fieraru}, \bibinfo{person}{Mihai Zanfir}, \bibinfo{person}{Teodor Szente}, \bibinfo{person}{Eduard Bazavan}, \bibinfo{person}{Vlad Olaru}, {and} \bibinfo{person}{Cristian Sminchisescu}.} \bibinfo{year}{2021}\natexlab{c}.
\newblock \showarticletitle{Remips: Physically consistent 3d reconstruction of multiple interacting people under weak supervision}.
\newblock \bibinfo{journal}{\emph{NeurIPS}}  \bibinfo{volume}{34} (\bibinfo{year}{2021}), \bibinfo{pages}{19385--19397}.
\newblock


\bibitem[Ghorbani et~al\mbox{.}(2020)]%
        {Movi}
\bibfield{author}{\bibinfo{person}{Saeed Ghorbani}, \bibinfo{person}{Kimia Mahdaviani}, \bibinfo{person}{Anne Thaler}, \bibinfo{person}{Konrad Kording}, \bibinfo{person}{Douglas~James Cook}, \bibinfo{person}{Gunnar Blohm}, {and} \bibinfo{person}{Nikolaus~F. Troje}.} \bibinfo{year}{2020}\natexlab{}.
\newblock \showarticletitle{{MoVi: A Large Multipurpose Motion and Video Dataset}}. \bibinfo{publisher}{Borealis}.
\newblock
\urldef\tempurl%
\url{https://doi.org/10.5683/SP2/JRHDRN}
\showDOI{\tempurl}


\bibitem[Guo et~al\mbox{.}(2022)]%
        {expi}
\bibfield{author}{\bibinfo{person}{Wen Guo}, \bibinfo{person}{Xiaoyu Bie}, \bibinfo{person}{Xavier Alameda-Pineda}, {and} \bibinfo{person}{Francesc Moreno-Noguer}.} \bibinfo{year}{2022}\natexlab{}.
\newblock \showarticletitle{Multi-person extreme motion prediction}. In \bibinfo{booktitle}{\emph{CVPR}}. \bibinfo{pages}{13053--13064}.
\newblock


\bibitem[Huang et~al\mbox{.}(2020)]%
        {dynamicmatching}
\bibfield{author}{\bibinfo{person}{Congzhentao Huang}, \bibinfo{person}{Shuai Jiang}, \bibinfo{person}{Yang Li}, \bibinfo{person}{Ziyue Zhang}, \bibinfo{person}{Jason Traish}, \bibinfo{person}{Chen Deng}, \bibinfo{person}{Sam Ferguson}, {and} \bibinfo{person}{Richard~Yi Da~Xu}.} \bibinfo{year}{2020}\natexlab{}.
\newblock \showarticletitle{End-to-end dynamic matching network for multi-view multi-person 3d pose estimation}. In \bibinfo{booktitle}{\emph{ECCV}}. Springer, \bibinfo{pages}{477--493}.
\newblock


\bibitem[Ionescu et~al\mbox{.}(2014)]%
        {h36m}
\bibfield{author}{\bibinfo{person}{Catalin Ionescu}, \bibinfo{person}{Dragos Papava}, \bibinfo{person}{Vlad Olaru}, {and} \bibinfo{person}{Cristian Sminchisescu}.} \bibinfo{year}{2014}\natexlab{}.
\newblock \showarticletitle{Human3.6M: Large Scale Datasets and Predictive Methods for 3D Human Sensing in Natural Environments}.
\newblock \bibinfo{journal}{\emph{IEEE Transactions on Pattern Analysis and Machine Intelligence}} \bibinfo{volume}{36}, \bibinfo{number}{7} (\bibinfo{year}{2014}), \bibinfo{pages}{1325--1339}.
\newblock


\bibitem[Iskakov et~al\mbox{.}(2019)]%
        {lrtri}
\bibfield{author}{\bibinfo{person}{Karim Iskakov}, \bibinfo{person}{Egor Burkov}, \bibinfo{person}{Victor Lempitsky}, {and} \bibinfo{person}{Yury Malkov}.} \bibinfo{year}{2019}\natexlab{}.
\newblock \showarticletitle{Learnable triangulation of human pose}. In \bibinfo{booktitle}{\emph{ICCV}}. \bibinfo{pages}{7718--7727}.
\newblock


\bibitem[Jocher(2020)]%
        {yolov5}
\bibfield{author}{\bibinfo{person}{Glenn Jocher}.} \bibinfo{year}{2020}\natexlab{}.
\newblock \bibinfo{booktitle}{\emph{Ultralytics YOLOv5}}.
\newblock
\urldef\tempurl%
\url{https://doi.org/10.5281/zenodo.3908559}
\showDOI{\tempurl}


\bibitem[Joo et~al\mbox{.}(2015)]%
        {panoptic}
\bibfield{author}{\bibinfo{person}{Hanbyul Joo}, \bibinfo{person}{Hao Liu}, \bibinfo{person}{Lei Tan}, \bibinfo{person}{Lin Gui}, \bibinfo{person}{Bart Nabbe}, \bibinfo{person}{Iain Matthews}, \bibinfo{person}{Takeo Kanade}, \bibinfo{person}{Shohei Nobuhara}, {and} \bibinfo{person}{Yaser Sheikh}.} \bibinfo{year}{2015}\natexlab{}.
\newblock \showarticletitle{Panoptic studio: A massively multiview system for social motion capture}. In \bibinfo{booktitle}{\emph{ICCV}}. \bibinfo{pages}{3334--3342}.
\newblock


\bibitem[Joo et~al\mbox{.}(2021)]%
        {eft}
\bibfield{author}{\bibinfo{person}{Hanbyul Joo}, \bibinfo{person}{Natalia Neverova}, {and} \bibinfo{person}{Andrea Vedaldi}.} \bibinfo{year}{2021}\natexlab{}.
\newblock \showarticletitle{Exemplar fine-tuning for 3d human model fitting towards in-the-wild 3d human pose estimation}. In \bibinfo{booktitle}{\emph{3DV}}. IEEE, \bibinfo{pages}{42--52}.
\newblock


\bibitem[Katircioglu et~al\mbox{.}(2021)]%
        {dyadicmotion}
\bibfield{author}{\bibinfo{person}{Isinsu Katircioglu}, \bibinfo{person}{Costa Georgantas}, \bibinfo{person}{Mathieu Salzmann}, {and} \bibinfo{person}{Pascal Fua}.} \bibinfo{year}{2021}\natexlab{}.
\newblock \showarticletitle{Dyadic human motion prediction}.
\newblock \bibinfo{journal}{\emph{arXiv preprint arXiv:2112.00396}} (\bibinfo{year}{2021}).
\newblock


\bibitem[Kingma and Ba(2015)]%
        {adam}
\bibfield{author}{\bibinfo{person}{Diederik~P. Kingma} {and} \bibinfo{person}{Jimmy Ba}.} \bibinfo{year}{2015}\natexlab{}.
\newblock \showarticletitle{Adam: {A} Method for Stochastic Optimization}. In \bibinfo{booktitle}{\emph{ICLR}}.
\newblock


\bibitem[Kolotouros et~al\mbox{.}(2019)]%
        {spin}
\bibfield{author}{\bibinfo{person}{Nikos Kolotouros}, \bibinfo{person}{Georgios Pavlakos}, \bibinfo{person}{Michael~J Black}, {and} \bibinfo{person}{Kostas Daniilidis}.} \bibinfo{year}{2019}\natexlab{}.
\newblock \showarticletitle{Learning to reconstruct 3D human pose and shape via model-fitting in the loop}. In \bibinfo{booktitle}{\emph{CVPR}}. \bibinfo{pages}{2252--2261}.
\newblock


\bibitem[Lin and Lee(2020)]%
        {hdnet}
\bibfield{author}{\bibinfo{person}{Jiahao Lin} {and} \bibinfo{person}{Gim~Hee Lee}.} \bibinfo{year}{2020}\natexlab{}.
\newblock \showarticletitle{Hdnet: Human depth estimation for multi-person camera-space localization}. In \bibinfo{booktitle}{\emph{ECCV}}. Springer, \bibinfo{pages}{633--648}.
\newblock


\bibitem[Lin and Lee(2021)]%
        {planesweep}
\bibfield{author}{\bibinfo{person}{Jiahao Lin} {and} \bibinfo{person}{Gim~Hee Lee}.} \bibinfo{year}{2021}\natexlab{}.
\newblock \showarticletitle{Multi-view multi-person 3d pose estimation with plane sweep stereo}. In \bibinfo{booktitle}{\emph{CVPR}}. \bibinfo{pages}{11886--11895}.
\newblock


\bibitem[Lin et~al\mbox{.}(2017)]%
        {focalloss}
\bibfield{author}{\bibinfo{person}{Tsung-Yi Lin}, \bibinfo{person}{Priya Goyal}, \bibinfo{person}{Ross Girshick}, \bibinfo{person}{Kaiming He}, {and} \bibinfo{person}{Piotr Doll{\'a}r}.} \bibinfo{year}{2017}\natexlab{}.
\newblock \showarticletitle{Focal loss for dense object detection}. In \bibinfo{booktitle}{\emph{ICCV}}. \bibinfo{pages}{2980--2988}.
\newblock


\bibitem[Lin et~al\mbox{.}(2014)]%
        {coco}
\bibfield{author}{\bibinfo{person}{Tsung-Yi Lin}, \bibinfo{person}{Michael Maire}, \bibinfo{person}{Serge Belongie}, \bibinfo{person}{James Hays}, \bibinfo{person}{Pietro Perona}, \bibinfo{person}{Deva Ramanan}, \bibinfo{person}{Piotr Doll{\'a}r}, {and} \bibinfo{person}{C~Lawrence Zitnick}.} \bibinfo{year}{2014}\natexlab{}.
\newblock \showarticletitle{Microsoft coco: Common objects in context}. In \bibinfo{booktitle}{\emph{ECCV}}. Springer, \bibinfo{pages}{740--755}.
\newblock


\bibitem[Liu et~al\mbox{.}(2021)]%
        {neuralactor}
\bibfield{author}{\bibinfo{person}{Lingjie Liu}, \bibinfo{person}{Marc Habermann}, \bibinfo{person}{Viktor Rudnev}, \bibinfo{person}{Kripasindhu Sarkar}, \bibinfo{person}{Jiatao Gu}, {and} \bibinfo{person}{Christian Theobalt}.} \bibinfo{year}{2021}\natexlab{}.
\newblock \showarticletitle{Neural actor: Neural free-view synthesis of human actors with pose control}.
\newblock \bibinfo{journal}{\emph{ACM Transactions on Graphics}} \bibinfo{volume}{40}, \bibinfo{number}{6} (\bibinfo{year}{2021}), \bibinfo{numpages}{16}~pages.
\newblock


\bibitem[Liu et~al\mbox{.}(2022)]%
        {dsed}
\bibfield{author}{\bibinfo{person}{Qihao Liu}, \bibinfo{person}{Yi Zhang}, \bibinfo{person}{Song Bai}, {and} \bibinfo{person}{Alan Yuille}.} \bibinfo{year}{2022}\natexlab{}.
\newblock \showarticletitle{Explicit Occlusion Reasoning for Multi-person 3D Human Pose Estimation}. In \bibinfo{booktitle}{\emph{ECCV}}. Springer, \bibinfo{pages}{497--517}.
\newblock


\bibitem[Loper et~al\mbox{.}(2015)]%
        {SMPL}
\bibfield{author}{\bibinfo{person}{Matthew Loper}, \bibinfo{person}{Naureen Mahmood}, \bibinfo{person}{Javier Romero}, \bibinfo{person}{Gerard Pons-Moll}, {and} \bibinfo{person}{Michael~J. Black}.} \bibinfo{year}{2015}\natexlab{}.
\newblock \showarticletitle{SMPL: A Skinned Multi-Person Linear Model}.
\newblock \bibinfo{journal}{\emph{ACM Transactions on Graphics}} \bibinfo{volume}{34}, \bibinfo{number}{6} (\bibinfo{date}{Nov} \bibinfo{year}{2015}), \bibinfo{numpages}{16}~pages.
\newblock


\bibitem[Mahmood et~al\mbox{.}(2019)]%
        {amass}
\bibfield{author}{\bibinfo{person}{Naureen Mahmood}, \bibinfo{person}{Nima Ghorbani}, \bibinfo{person}{Nikolaus~F Troje}, \bibinfo{person}{Gerard Pons-Moll}, {and} \bibinfo{person}{Michael~J Black}.} \bibinfo{year}{2019}\natexlab{}.
\newblock \showarticletitle{AMASS: Archive of motion capture as surface shapes}. In \bibinfo{booktitle}{\emph{ICCV}}. \bibinfo{pages}{5442--5451}.
\newblock


\bibitem[Mehta et~al\mbox{.}(2017)]%
        {mono-3dhp2017}
\bibfield{author}{\bibinfo{person}{Dushyant Mehta}, \bibinfo{person}{Helge Rhodin}, \bibinfo{person}{Dan Casas}, \bibinfo{person}{Pascal Fua}, \bibinfo{person}{Oleksandr Sotnychenko}, \bibinfo{person}{Weipeng Xu}, {and} \bibinfo{person}{Christian Theobalt}.} \bibinfo{year}{2017}\natexlab{}.
\newblock \showarticletitle{Monocular 3D Human Pose Estimation In The Wild Using Improved CNN Supervision}. In \bibinfo{booktitle}{\emph{3DV}}. IEEE.
\newblock
\urldef\tempurl%
\url{http://gvv.mpi-inf.mpg.de/3dhp_dataset}
\showURL{%
\tempurl}


\bibitem[Mehta et~al\mbox{.}(2020)]%
        {xnect}
\bibfield{author}{\bibinfo{person}{Dushyant Mehta}, \bibinfo{person}{Oleksandr Sotnychenko}, \bibinfo{person}{Franziska Mueller}, \bibinfo{person}{Weipeng Xu}, \bibinfo{person}{Mohamed Elgharib}, \bibinfo{person}{Pascal Fua}, \bibinfo{person}{Hans-Peter Seidel}, \bibinfo{person}{Helge Rhodin}, \bibinfo{person}{Gerard Pons-Moll}, {and} \bibinfo{person}{Christian Theobalt}.} \bibinfo{year}{2020}\natexlab{}.
\newblock \showarticletitle{{XNect}: Real-time Multi-Person {3D} Motion Capture with a Single {RGB} Camera}.
\newblock \bibinfo{journal}{\emph{ACM Transactions on Graphics}} \bibinfo{volume}{39}, \bibinfo{number}{4} (\bibinfo{date}{July} \bibinfo{year}{2020}), \bibinfo{numpages}{17}~pages.
\newblock


\bibitem[Mehta et~al\mbox{.}(2018)]%
        {muco}
\bibfield{author}{\bibinfo{person}{Dushyant Mehta}, \bibinfo{person}{Oleksandr Sotnychenko}, \bibinfo{person}{Franziska Mueller}, \bibinfo{person}{Weipeng Xu}, \bibinfo{person}{Srinath Sridhar}, \bibinfo{person}{Gerard Pons-Moll}, {and} \bibinfo{person}{Christian Theobalt}.} \bibinfo{year}{2018}\natexlab{}.
\newblock \showarticletitle{Single-shot multi-person 3d pose estimation from monocular rgb}. In \bibinfo{booktitle}{\emph{3DV}}. \bibinfo{pages}{120--130}.
\newblock


\bibitem[Mihajlovic et~al\mbox{.}(2022)]%
        {keypointnerf}
\bibfield{author}{\bibinfo{person}{Marko Mihajlovic}, \bibinfo{person}{Aayush Bansal}, \bibinfo{person}{Michael Zollhoefer}, \bibinfo{person}{Siyu Tang}, {and} \bibinfo{person}{Shunsuke Saito}.} \bibinfo{year}{2022}\natexlab{}.
\newblock \showarticletitle{KeypointNeRF: Generalizing image-based volumetric avatars using relative spatial encoding of keypoints}. In \bibinfo{booktitle}{\emph{ECCV}}. Springer, \bibinfo{pages}{179--197}.
\newblock


\bibitem[Mildenhall et~al\mbox{.}(2021)]%
        {nerf}
\bibfield{author}{\bibinfo{person}{Ben Mildenhall}, \bibinfo{person}{Pratul~P Srinivasan}, \bibinfo{person}{Matthew Tancik}, \bibinfo{person}{Jonathan~T Barron}, \bibinfo{person}{Ravi Ramamoorthi}, {and} \bibinfo{person}{Ren Ng}.} \bibinfo{year}{2021}\natexlab{}.
\newblock \showarticletitle{Nerf: Representing scenes as neural radiance fields for view synthesis}.
\newblock \bibinfo{journal}{\emph{Commun. ACM}} \bibinfo{volume}{65}, \bibinfo{number}{1} (\bibinfo{year}{2021}), \bibinfo{pages}{99--106}.
\newblock


\bibitem[Moon et~al\mbox{.}(2019)]%
        {ROOTNET}
\bibfield{author}{\bibinfo{person}{Gyeongsik Moon}, \bibinfo{person}{Juyong Chang}, {and} \bibinfo{person}{Kyoung~Mu Lee}.} \bibinfo{year}{2019}\natexlab{}.
\newblock \showarticletitle{Camera Distance-aware Top-down Approach for 3D Multi-person Pose Estimation from a Single RGB Image}. In \bibinfo{booktitle}{\emph{ICCV}}. \bibinfo{pages}{10133--10142}.
\newblock


\bibitem[Ofli et~al\mbox{.}(2013)]%
        {mhad}
\bibfield{author}{\bibinfo{person}{Ferda Ofli}, \bibinfo{person}{Rizwan Chaudhry}, \bibinfo{person}{Gregorij Kurillo}, \bibinfo{person}{Ren{\'e} Vidal}, {and} \bibinfo{person}{Ruzena Bajcsy}.} \bibinfo{year}{2013}\natexlab{}.
\newblock \showarticletitle{Berkeley mhad: A comprehensive multimodal human action database}. In \bibinfo{booktitle}{\emph{WACV}}. \bibinfo{pages}{53--60}.
\newblock


\bibitem[Patel et~al\mbox{.}(2021)]%
        {agora}
\bibfield{author}{\bibinfo{person}{Priyanka Patel}, \bibinfo{person}{Chun-Hao~P Huang}, \bibinfo{person}{Joachim Tesch}, \bibinfo{person}{David~T Hoffmann}, \bibinfo{person}{Shashank Tripathi}, {and} \bibinfo{person}{Michael~J Black}.} \bibinfo{year}{2021}\natexlab{}.
\newblock \showarticletitle{AGORA: Avatars in geography optimized for regression analysis}. In \bibinfo{booktitle}{\emph{CVPR}}. \bibinfo{pages}{13468--13478}.
\newblock


\bibitem[Pavlakos et~al\mbox{.}(2019)]%
        {SMPL-X}
\bibfield{author}{\bibinfo{person}{Georgios Pavlakos}, \bibinfo{person}{Vasileios Choutas}, \bibinfo{person}{Nima Ghorbani}, \bibinfo{person}{Timo Bolkart}, \bibinfo{person}{Ahmed A.~A. Osman}, \bibinfo{person}{Dimitrios Tzionas}, {and} \bibinfo{person}{Michael~J. Black}.} \bibinfo{year}{2019}\natexlab{}.
\newblock \showarticletitle{Expressive Body Capture: {3D} Hands, Face, and Body from a Single Image}. In \bibinfo{booktitle}{\emph{CVPR}}. \bibinfo{pages}{10975--10985}.
\newblock


\bibitem[Peng et~al\mbox{.}(2021)]%
        {neuralbody}
\bibfield{author}{\bibinfo{person}{Sida Peng}, \bibinfo{person}{Yuanqing Zhang}, \bibinfo{person}{Yinghao Xu}, \bibinfo{person}{Qianqian Wang}, \bibinfo{person}{Qing Shuai}, \bibinfo{person}{Hujun Bao}, {and} \bibinfo{person}{Xiaowei Zhou}.} \bibinfo{year}{2021}\natexlab{}.
\newblock \showarticletitle{Neural body: Implicit neural representations with structured latent codes for novel view synthesis of dynamic humans}. In \bibinfo{booktitle}{\emph{CVPR}}. \bibinfo{pages}{9054--9063}.
\newblock


\bibitem[Qiu et~al\mbox{.}(2023)]%
        {psvt}
\bibfield{author}{\bibinfo{person}{Zhongwei Qiu}, \bibinfo{person}{Yang Qiansheng}, \bibinfo{person}{Jian Wang}, \bibinfo{person}{Haocheng Feng}, \bibinfo{person}{Junyu Han}, \bibinfo{person}{Errui Ding}, \bibinfo{person}{Chang Xu}, \bibinfo{person}{Dongmei Fu}, {and} \bibinfo{person}{Jingdong Wang}.} \bibinfo{year}{2023}\natexlab{}.
\newblock \showarticletitle{PSVT: End-to-End Multi-person 3D Pose and Shape Estimation with Progressive Video Transformers}. In \bibinfo{booktitle}{\emph{CVPR}}.
\newblock


\bibitem[Rempe et~al\mbox{.}(2021)]%
        {humor}
\bibfield{author}{\bibinfo{person}{Davis Rempe}, \bibinfo{person}{Tolga Birdal}, \bibinfo{person}{Aaron Hertzmann}, \bibinfo{person}{Jimei Yang}, \bibinfo{person}{Srinath Sridhar}, {and} \bibinfo{person}{Leonidas~J Guibas}.} \bibinfo{year}{2021}\natexlab{}.
\newblock \showarticletitle{Humor: 3d human motion model for robust pose estimation}. In \bibinfo{booktitle}{\emph{ICCV}}. \bibinfo{pages}{11488--11499}.
\newblock


\bibitem[Robinette et~al\mbox{.}(2002)]%
        {CAESAR}
\bibfield{author}{\bibinfo{person}{Kathleen~M Robinette}, \bibinfo{person}{Sherri Blackwell}, \bibinfo{person}{Hein Daanen}, \bibinfo{person}{Mark Boehmer}, \bibinfo{person}{Scott Fleming}, \bibinfo{person}{Tina Brill}, \bibinfo{person}{David Hoeferlin}, {and} \bibinfo{person}{Dennis Burnsides}.} \bibinfo{year}{2002}\natexlab{}.
\newblock \showarticletitle{Civilian American and European surface anthropometry resource (CAESAR), final report, volume I: Summary}.
\newblock \bibinfo{journal}{\emph{Sytronics Inc Dayton Oh}} (\bibinfo{year}{2002}).
\newblock


\bibitem[Shuai et~al\mbox{.}(2022)]%
        {multinb}
\bibfield{author}{\bibinfo{person}{Qing Shuai}, \bibinfo{person}{Chen Geng}, \bibinfo{person}{Qi Fang}, \bibinfo{person}{Sida Peng}, \bibinfo{person}{Wenhao Shen}, \bibinfo{person}{Xiaowei Zhou}, {and} \bibinfo{person}{Hujun Bao}.} \bibinfo{year}{2022}\natexlab{}.
\newblock \showarticletitle{Novel view synthesis of human interactions from sparse multi-view videos}. In \bibinfo{booktitle}{\emph{SIGGRAPH}}. \bibinfo{pages}{1--10}.
\newblock


\bibitem[Sigal et~al\mbox{.}(2010)]%
        {humaneva}
\bibfield{author}{\bibinfo{person}{Leonid Sigal}, \bibinfo{person}{Alexandru~O Balan}, {and} \bibinfo{person}{Michael~J Black}.} \bibinfo{year}{2010}\natexlab{}.
\newblock \showarticletitle{Humaneva: Synchronized video and motion capture dataset and baseline algorithm for evaluation of articulated human motion}.
\newblock \bibinfo{journal}{\emph{International Journal of Computer Vision}} \bibinfo{volume}{87}, \bibinfo{number}{1-2} (\bibinfo{year}{2010}), \bibinfo{pages}{4}.
\newblock


\bibitem[Su et~al\mbox{.}(2022)]%
        {virtualpose}
\bibfield{author}{\bibinfo{person}{Jiajun Su}, \bibinfo{person}{Chunyu Wang}, \bibinfo{person}{Xiaoxuan Ma}, \bibinfo{person}{Wenjun Zeng}, {and} \bibinfo{person}{Yizhou Wang}.} \bibinfo{year}{2022}\natexlab{}.
\newblock \showarticletitle{VirtualPose: Learning Generalizable 3D Human Pose Models from Virtual Data}. In \bibinfo{booktitle}{\emph{ECCV}}. Springer, \bibinfo{pages}{55--71}.
\newblock


\bibitem[Sun et~al\mbox{.}(2022)]%
        {putting}
\bibfield{author}{\bibinfo{person}{Yu Sun}, \bibinfo{person}{Wu Liu}, \bibinfo{person}{Qian Bao}, \bibinfo{person}{Yili Fu}, \bibinfo{person}{Tao Mei}, {and} \bibinfo{person}{Michael~J Black}.} \bibinfo{year}{2022}\natexlab{}.
\newblock \showarticletitle{Putting people in their place: Monocular regression of 3d people in depth}. In \bibinfo{booktitle}{\emph{CVPR}}. \bibinfo{pages}{13243--13252}.
\newblock


\bibitem[Trumble et~al\mbox{.}(2017)]%
        {totalcapture}
\bibfield{author}{\bibinfo{person}{Matt Trumble}, \bibinfo{person}{Andrew Gilbert}, \bibinfo{person}{Charles Malleson}, \bibinfo{person}{Adrian Hilton}, {and} \bibinfo{person}{John Collomosse}.} \bibinfo{year}{2017}\natexlab{}.
\newblock \showarticletitle{Total Capture: 3D Human Pose Estimation Fusing Video and Inertial Sensors}. In \bibinfo{booktitle}{\emph{BMVC}}.
\newblock


\bibitem[Tu et~al\mbox{.}(2020)]%
        {voxelpose}
\bibfield{author}{\bibinfo{person}{Hanyue Tu}, \bibinfo{person}{Chunyu Wang}, {and} \bibinfo{person}{Wenjun Zeng}.} \bibinfo{year}{2020}\natexlab{}.
\newblock \showarticletitle{Voxelpose: Towards multi-camera 3d human pose estimation in wild environment}. In \bibinfo{booktitle}{\emph{ECCV}}. Springer, \bibinfo{pages}{197--212}.
\newblock


\bibitem[Varol et~al\mbox{.}(2017)]%
        {surreal}
\bibfield{author}{\bibinfo{person}{Gul Varol}, \bibinfo{person}{Javier Romero}, \bibinfo{person}{Xavier Martin}, \bibinfo{person}{Naureen Mahmood}, \bibinfo{person}{Michael~J Black}, \bibinfo{person}{Ivan Laptev}, {and} \bibinfo{person}{Cordelia Schmid}.} \bibinfo{year}{2017}\natexlab{}.
\newblock \showarticletitle{Learning from synthetic humans}. In \bibinfo{booktitle}{\emph{CVPR}}. \bibinfo{pages}{109--117}.
\newblock


\bibitem[Von~Marcard et~al\mbox{.}(2018)]%
        {3DPW}
\bibfield{author}{\bibinfo{person}{Timo Von~Marcard}, \bibinfo{person}{Roberto Henschel}, \bibinfo{person}{Michael~J Black}, \bibinfo{person}{Bodo Rosenhahn}, {and} \bibinfo{person}{Gerard Pons-Moll}.} \bibinfo{year}{2018}\natexlab{}.
\newblock \showarticletitle{Recovering accurate 3d human pose in the wild using imus and a moving camera}. In \bibinfo{booktitle}{\emph{ECCV}}. \bibinfo{pages}{601--617}.
\newblock


\bibitem[Wandt et~al\mbox{.}(2021)]%
        {canonpose}
\bibfield{author}{\bibinfo{person}{Bastian Wandt}, \bibinfo{person}{Marco Rudolph}, \bibinfo{person}{Petrissa Zell}, \bibinfo{person}{Helge Rhodin}, {and} \bibinfo{person}{Bodo Rosenhahn}.} \bibinfo{year}{2021}\natexlab{}.
\newblock \showarticletitle{CanonPose: Self-Supervised Monocular 3D Human Pose Estimation in the Wild}. In \bibinfo{booktitle}{\emph{CVPR}}.
\newblock


\bibitem[Wang et~al\mbox{.}(2020a)]%
        {hmor}
\bibfield{author}{\bibinfo{person}{Can Wang}, \bibinfo{person}{Jiefeng Li}, \bibinfo{person}{Wentao Liu}, \bibinfo{person}{Chen Qian}, {and} \bibinfo{person}{Cewu Lu}.} \bibinfo{year}{2020}\natexlab{a}.
\newblock \showarticletitle{Hmor: Hierarchical multi-person ordinal relations for monocular multi-person 3d pose estimation}. In \bibinfo{booktitle}{\emph{ECCV}}. Springer, \bibinfo{pages}{242--259}.
\newblock


\bibitem[Wang et~al\mbox{.}(2020b)]%
        {hrnet}
\bibfield{author}{\bibinfo{person}{Jingdong Wang}, \bibinfo{person}{Ke Sun}, \bibinfo{person}{Tianheng Cheng}, \bibinfo{person}{Borui Jiang}, \bibinfo{person}{Chaorui Deng}, \bibinfo{person}{Yang Zhao}, \bibinfo{person}{Dong Liu}, \bibinfo{person}{Yadong Mu}, \bibinfo{person}{Mingkui Tan}, \bibinfo{person}{Xinggang Wang}, {et~al\mbox{.}}} \bibinfo{year}{2020}\natexlab{b}.
\newblock \showarticletitle{Deep high-resolution representation learning for visual recognition}.
\newblock \bibinfo{journal}{\emph{IEEE Transactions on Pattern Analysis and Machine Intelligence}} \bibinfo{volume}{43}, \bibinfo{number}{10} (\bibinfo{year}{2020}), \bibinfo{pages}{3349--3364}.
\newblock


\bibitem[Wang et~al\mbox{.}(2021)]%
        {direct}
\bibfield{author}{\bibinfo{person}{Tao Wang}, \bibinfo{person}{Jianfeng Zhang}, \bibinfo{person}{Yujun Cai}, \bibinfo{person}{Shuicheng Yan}, {and} \bibinfo{person}{Jiashi Feng}.} \bibinfo{year}{2021}\natexlab{}.
\newblock \showarticletitle{Direct Multi-view Multi-person 3D Human Pose Estimation}.
\newblock \bibinfo{journal}{\emph{NeurIPS}}  \bibinfo{volume}{34} (\bibinfo{year}{2021}), \bibinfo{pages}{13153--13164}.
\newblock


\bibitem[Wang et~al\mbox{.}(2022)]%
        {das}
\bibfield{author}{\bibinfo{person}{Zitian Wang}, \bibinfo{person}{Xuecheng Nie}, \bibinfo{person}{Xiaochao Qu}, \bibinfo{person}{Yunpeng Chen}, {and} \bibinfo{person}{Si Liu}.} \bibinfo{year}{2022}\natexlab{}.
\newblock \showarticletitle{Distribution-aware single-stage models for multi-person 3D pose estimation}. In \bibinfo{booktitle}{\emph{CVPR}}. \bibinfo{pages}{13096--13105}.
\newblock


\bibitem[Weng et~al\mbox{.}(2022)]%
        {humannerf}
\bibfield{author}{\bibinfo{person}{Chung-Yi Weng}, \bibinfo{person}{Brian Curless}, \bibinfo{person}{Pratul~P Srinivasan}, \bibinfo{person}{Jonathan~T Barron}, {and} \bibinfo{person}{Ira Kemelmacher-Shlizerman}.} \bibinfo{year}{2022}\natexlab{}.
\newblock \showarticletitle{Humannerf: Free-viewpoint rendering of moving people from monocular video}. In \bibinfo{booktitle}{\emph{CVPR}}. \bibinfo{pages}{16210--16220}.
\newblock


\bibitem[Wu et~al\mbox{.}(2021)]%
        {graph}
\bibfield{author}{\bibinfo{person}{Size Wu}, \bibinfo{person}{Sheng Jin}, \bibinfo{person}{Wentao Liu}, \bibinfo{person}{Lei Bai}, \bibinfo{person}{Chen Qian}, \bibinfo{person}{Dong Liu}, {and} \bibinfo{person}{Wanli Ouyang}.} \bibinfo{year}{2021}\natexlab{}.
\newblock \showarticletitle{Graph-based 3d multi-person pose estimation using multi-view images}. In \bibinfo{booktitle}{\emph{ICCV}}. \bibinfo{pages}{11148--11157}.
\newblock


\bibitem[Xu et~al\mbox{.}(2020)]%
        {ghum}
\bibfield{author}{\bibinfo{person}{Hongyi Xu}, \bibinfo{person}{Eduard~Gabriel Bazavan}, \bibinfo{person}{Andrei Zanfir}, \bibinfo{person}{William~T Freeman}, \bibinfo{person}{Rahul Sukthankar}, {and} \bibinfo{person}{Cristian Sminchisescu}.} \bibinfo{year}{2020}\natexlab{}.
\newblock \showarticletitle{GHUM \& GHUML: Generative 3D Human Shape and Articulated Pose Models}. In \bibinfo{booktitle}{\emph{CVPR}}. \bibinfo{pages}{6184--6193}.
\newblock


\bibitem[Ye et~al\mbox{.}(2022)]%
        {fastervoxelpose}
\bibfield{author}{\bibinfo{person}{Hang Ye}, \bibinfo{person}{Wentao Zhu}, \bibinfo{person}{Chunyu Wang}, \bibinfo{person}{Rujie Wu}, {and} \bibinfo{person}{Yizhou Wang}.} \bibinfo{year}{2022}\natexlab{}.
\newblock \showarticletitle{Faster VoxelPose: Real-time 3D Human Pose Estimation by Orthographic Projection}. In \bibinfo{booktitle}{\emph{ECCV}}. Springer, \bibinfo{pages}{142--159}.
\newblock


\bibitem[Ye et~al\mbox{.}(2023)]%
        {slahmr}
\bibfield{author}{\bibinfo{person}{Vickie Ye}, \bibinfo{person}{Georgios Pavlakos}, \bibinfo{person}{Jitendra Malik}, {and} \bibinfo{person}{Angjoo Kanazawa}.} \bibinfo{year}{2023}\natexlab{}.
\newblock \showarticletitle{Decoupling Human and Camera Motion from Videos in the Wild}. In \bibinfo{booktitle}{\emph{CVPR}}.
\newblock


\bibitem[Yin et~al\mbox{.}(2023)]%
        {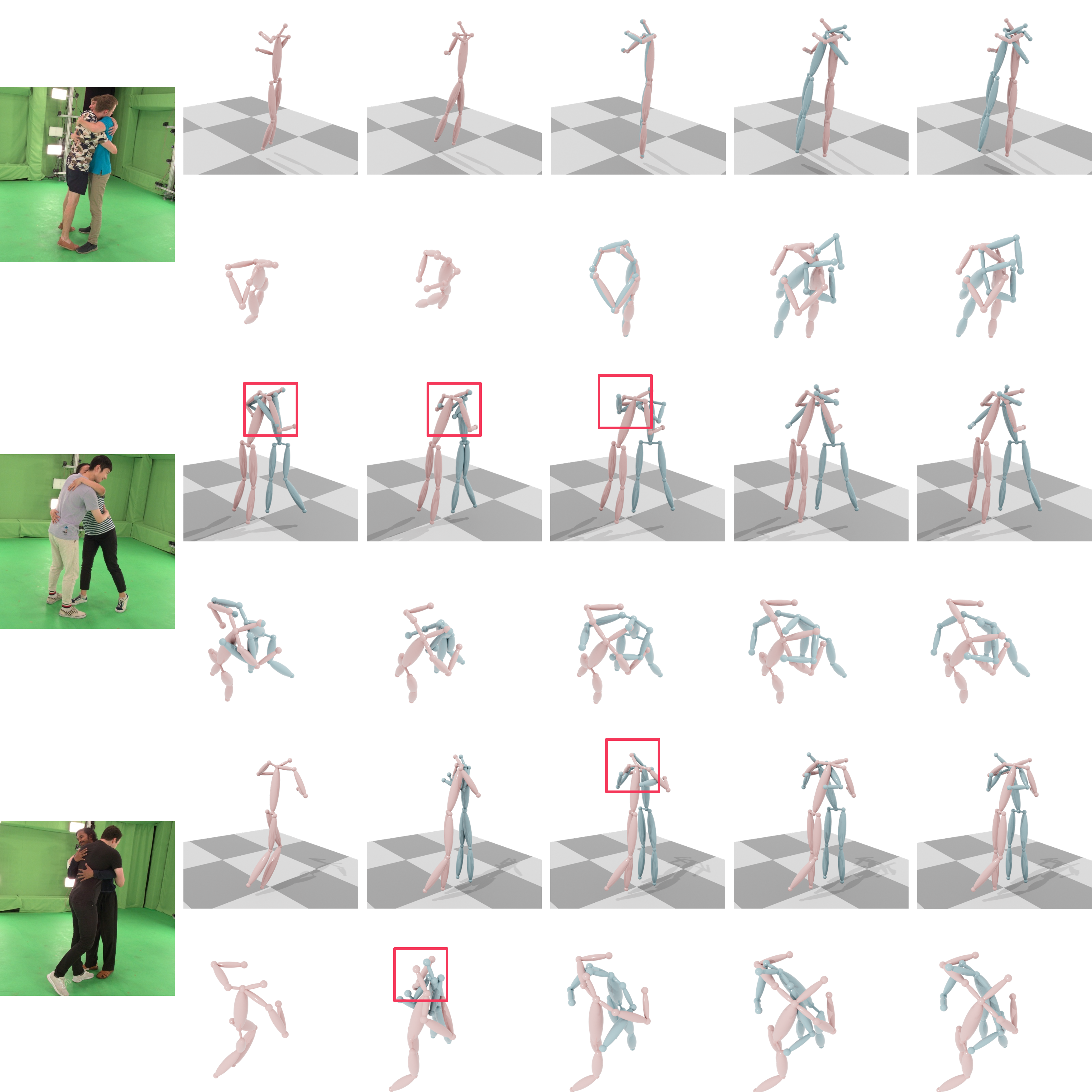}
\bibfield{author}{\bibinfo{person}{Yifei Yin}, \bibinfo{person}{Chen Guo}, \bibinfo{person}{Manuel Kaufmann}, \bibinfo{person}{Juan Zarate}, \bibinfo{person}{Jie Song}, {and} \bibinfo{person}{Otmar Hilliges}.} \bibinfo{year}{2023}\natexlab{}.
\newblock \showarticletitle{Hi4D: 4D Instance Segmentation of Close Human Interaction}. In \bibinfo{booktitle}{\emph{CVPR}}. \bibinfo{pages}{17016--17027}.
\newblock


\bibitem[Yoon et~al\mbox{.}(2021)]%
        {humbi}
\bibfield{author}{\bibinfo{person}{Jae~Shin Yoon}, \bibinfo{person}{Zhixuan Yu}, \bibinfo{person}{Jaesik Park}, {and} \bibinfo{person}{Hyun~Soo Park}.} \bibinfo{year}{2021}\natexlab{}.
\newblock \showarticletitle{Humbi: A large multiview dataset of human body expressions and benchmark challenge}.
\newblock \bibinfo{journal}{\emph{IEEE Transactions on Pattern Analysis and Machine Intelligence}} \bibinfo{volume}{45}, \bibinfo{number}{1} (\bibinfo{year}{2021}), \bibinfo{pages}{623--640}.
\newblock


\bibitem[Yuan et~al\mbox{.}(2022)]%
        {glamr}
\bibfield{author}{\bibinfo{person}{Ye Yuan}, \bibinfo{person}{Umar Iqbal}, \bibinfo{person}{Pavlo Molchanov}, \bibinfo{person}{Kris Kitani}, {and} \bibinfo{person}{Jan Kautz}.} \bibinfo{year}{2022}\natexlab{}.
\newblock \showarticletitle{GLAMR: Global occlusion-aware human mesh recovery with dynamic cameras}. In \bibinfo{booktitle}{\emph{CVPR}}. \bibinfo{pages}{11038--11049}.
\newblock


\bibitem[Zhang et~al\mbox{.}(2020)]%
        {4d}
\bibfield{author}{\bibinfo{person}{Yuxiang Zhang}, \bibinfo{person}{Liang An}, \bibinfo{person}{Tao Yu}, \bibinfo{person}{Xiu Li}, \bibinfo{person}{Kun Li}, {and} \bibinfo{person}{Yebin Liu}.} \bibinfo{year}{2020}\natexlab{}.
\newblock \showarticletitle{4D association graph for realtime multi-person motion capture using multiple video cameras}. In \bibinfo{booktitle}{\emph{CVPR}}. \bibinfo{pages}{1324--1333}.
\newblock


\bibitem[Zhen et~al\mbox{.}(2020)]%
        {smap}
\bibfield{author}{\bibinfo{person}{Jianan Zhen}, \bibinfo{person}{Qi Fang}, \bibinfo{person}{Jiaming Sun}, \bibinfo{person}{Wentao Liu}, \bibinfo{person}{Wei Jiang}, \bibinfo{person}{Hujun Bao}, {and} \bibinfo{person}{Xiaowei Zhou}.} \bibinfo{year}{2020}\natexlab{}.
\newblock \showarticletitle{Smap: Single-shot multi-person absolute 3d pose estimation}. In \bibinfo{booktitle}{\emph{ECCV}}. Springer, \bibinfo{pages}{550--566}.
\newblock


\bibitem[Zhou et~al\mbox{.}(2022)]%
        {quickpose}
\bibfield{author}{\bibinfo{person}{Zhize Zhou}, \bibinfo{person}{Qing Shuai}, \bibinfo{person}{Yize Wang}, \bibinfo{person}{Qi Fang}, \bibinfo{person}{Xiaopeng Ji}, \bibinfo{person}{Fashuai Li}, \bibinfo{person}{Hujun Bao}, {and} \bibinfo{person}{Xiaowei Zhou}.} \bibinfo{year}{2022}\natexlab{}.
\newblock \showarticletitle{QuickPose: Real-time Multi-view Multi-person Pose Estimation in Crowded Scenes}. In \bibinfo{booktitle}{\emph{SIGGRAPH}}. \bibinfo{pages}{1--9}.
\newblock


\end{thebibliography}



\begin{thebibliography}{21}


\ifx \showCODEN    \undefined \def \showCODEN     #1{\unskip}     \fi
\ifx \showDOI      \undefined \def \showDOI       #1{#1}\fi
\ifx \showISBNx    \undefined \def \showISBNx     #1{\unskip}     \fi
\ifx \showISBNxiii \undefined \def \showISBNxiii  #1{\unskip}     \fi
\ifx \showISSN     \undefined \def \showISSN      #1{\unskip}     \fi
\ifx \showLCCN     \undefined \def \showLCCN      #1{\unskip}     \fi
\ifx \shownote     \undefined \def \shownote      #1{#1}          \fi
\ifx \showarticletitle \undefined \def \showarticletitle #1{#1}   \fi
\ifx \showURL      \undefined \def \showURL       {\relax}        \fi
\providecommand\bibfield[2]{#2}
\providecommand\bibinfo[2]{#2}
\providecommand\natexlab[1]{#1}
\providecommand\showeprint[2][]{arXiv:#2}

\bibitem[Belagiannis et~al\mbox{.}(2014)]%
        {belagiannis20143d}
\bibfield{author}{\bibinfo{person}{Vasileios Belagiannis}, \bibinfo{person}{Sikandar Amin}, \bibinfo{person}{Mykhaylo Andriluka}, \bibinfo{person}{Bernt Schiele}, \bibinfo{person}{Nassir Navab}, {and} \bibinfo{person}{Slobodan Ilic}.} \bibinfo{year}{2014}\natexlab{}.
\newblock \showarticletitle{3D pictorial structures for multiple human pose estimation}. In \bibinfo{booktitle}{\emph{Proceedings of the IEEE conference on computer vision and pattern recognition}}. \bibinfo{pages}{1669--1676}.
\newblock


\bibitem[Belagiannis et~al\mbox{.}(2015)]%
        {belagiannis20153d}
\bibfield{author}{\bibinfo{person}{Vasileios Belagiannis}, \bibinfo{person}{Sikandar Amin}, \bibinfo{person}{Mykhaylo Andriluka}, \bibinfo{person}{Bernt Schiele}, \bibinfo{person}{Nassir Navab}, {and} \bibinfo{person}{Slobodan Ilic}.} \bibinfo{year}{2015}\natexlab{}.
\newblock \showarticletitle{3d pictorial structures revisited: Multiple human pose estimation}.
\newblock \bibinfo{journal}{\emph{IEEE transactions on pattern analysis and machine intelligence}} \bibinfo{volume}{38}, \bibinfo{number}{10} (\bibinfo{year}{2015}), \bibinfo{pages}{1929--1942}.
\newblock


\bibitem[Dong et~al\mbox{.}(2019)]%
        {dong2019fast}
\bibfield{author}{\bibinfo{person}{Junting Dong}, \bibinfo{person}{Wen Jiang}, \bibinfo{person}{Qixing Huang}, \bibinfo{person}{Hujun Bao}, {and} \bibinfo{person}{Xiaowei Zhou}.} \bibinfo{year}{2019}\natexlab{}.
\newblock \showarticletitle{Fast and Robust Multi-Person 3D Pose Estimation from Multiple Views}.
\newblock \bibinfo{journal}{\emph{CVPR}}.
\newblock


\bibitem[Dong et~al\mbox{.}(2021)]%
        {dong2021shape}
\bibfield{author}{\bibinfo{person}{Zijian Dong}, \bibinfo{person}{Jie Song}, \bibinfo{person}{Xu Chen}, \bibinfo{person}{Chen Guo}, {and} \bibinfo{person}{Otmar Hilliges}.} \bibinfo{year}{2021}\natexlab{}.
\newblock \showarticletitle{Shape-aware multi-person pose estimation from multi-view images}. In \bibinfo{booktitle}{\emph{Proceedings of the IEEE/CVF International Conference on Computer Vision}}. \bibinfo{pages}{11158--11168}.
\newblock


\bibitem[Fang et~al\mbox{.}(2021)]%
        {Fang_2021_CVPR}
\bibfield{author}{\bibinfo{person}{Qi Fang}, \bibinfo{person}{Qing Shuai}, \bibinfo{person}{Junting Dong}, \bibinfo{person}{Hujun Bao}, {and} \bibinfo{person}{Xiaowei Zhou}.} \bibinfo{year}{2021}\natexlab{}.
\newblock \showarticletitle{Reconstructing 3D Human Pose by Watching Humans in the Mirror}. In \bibinfo{booktitle}{\emph{Proceedings of the IEEE/CVF Conference on Computer Vision and Pattern Recognition (CVPR)}}. \bibinfo{pages}{12814--12823}.
\newblock


\bibitem[Fieraru et~al\mbox{.}(2020)]%
        {Fieraru_2020_CVPR}
\bibfield{author}{\bibinfo{person}{Mihai Fieraru}, \bibinfo{person}{Mihai Zanfir}, \bibinfo{person}{Elisabeta Oneata}, \bibinfo{person}{Alin-Ionut Popa}, \bibinfo{person}{Vlad Olaru}, {and} \bibinfo{person}{Cristian Sminchisescu}.} \bibinfo{year}{2020}\natexlab{}.
\newblock \showarticletitle{Three-Dimensional Reconstruction of Human Interactions}. In \bibinfo{booktitle}{\emph{The IEEE/CVF Conference on Computer Vision and Pattern Recognition (CVPR)}}.
\newblock


\bibitem[Huang et~al\mbox{.}(2020)]%
        {huang2020end}
\bibfield{author}{\bibinfo{person}{Congzhentao Huang}, \bibinfo{person}{Shuai Jiang}, \bibinfo{person}{Yang Li}, \bibinfo{person}{Ziyue Zhang}, \bibinfo{person}{Jason Traish}, \bibinfo{person}{Chen Deng}, \bibinfo{person}{Sam Ferguson}, {and} \bibinfo{person}{Richard~Yi Da~Xu}.} \bibinfo{year}{2020}\natexlab{}.
\newblock \showarticletitle{End-to-end dynamic matching network for multi-view multi-person 3d pose estimation}. In \bibinfo{booktitle}{\emph{Computer Vision--ECCV 2020: 16th European Conference, Glasgow, UK, August 23--28, 2020, Proceedings, Part XXVIII 16}}. Springer, \bibinfo{pages}{477--493}.
\newblock


\bibitem[Jin et~al\mbox{.}(2022)]%
        {jin2022single}
\bibfield{author}{\bibinfo{person}{Lei Jin}, \bibinfo{person}{Chenyang Xu}, \bibinfo{person}{Xiaojuan Wang}, \bibinfo{person}{Yabo Xiao}, \bibinfo{person}{Yandong Guo}, \bibinfo{person}{Xuecheng Nie}, {and} \bibinfo{person}{Jian Zhao}.} \bibinfo{year}{2022}\natexlab{}.
\newblock \showarticletitle{Single-stage is enough: Multi-person absolute 3D pose estimation}. In \bibinfo{booktitle}{\emph{Proceedings of the IEEE/CVF Conference on Computer Vision and Pattern Recognition}}. \bibinfo{pages}{13086--13095}.
\newblock


\bibitem[Lin and Lee(2021)]%
        {lin2021sweepplane}
\bibfield{author}{\bibinfo{person}{Jiahao Lin} {and} \bibinfo{person}{Gim~Hee Lee}.} \bibinfo{year}{2021}\natexlab{}.
\newblock \showarticletitle{Multi-view multi-person 3d pose estimation with plane sweep stereo}. In \bibinfo{booktitle}{\emph{Proceedings of the IEEE/CVF Conference on Computer Vision and Pattern Recognition}}. \bibinfo{pages}{11886--11895}.
\newblock


\bibitem[Liu et~al\mbox{.}(2022)]%
        {liu2022explicit}
\bibfield{author}{\bibinfo{person}{Qihao Liu}, \bibinfo{person}{Yi Zhang}, \bibinfo{person}{Song Bai}, {and} \bibinfo{person}{Alan Yuille}.} \bibinfo{year}{2022}\natexlab{}.
\newblock \showarticletitle{Explicit Occlusion Reasoning for Multi-person 3D Human Pose Estimation}. In \bibinfo{booktitle}{\emph{Computer Vision--ECCV 2022: 17th European Conference, Tel Aviv, Israel, October 23--27, 2022, Proceedings, Part V}}. Springer, \bibinfo{pages}{497--517}.
\newblock


\bibitem[Qiu et~al\mbox{.}(2022)]%
        {qiu2022dynamic}
\bibfield{author}{\bibinfo{person}{Zhongwei Qiu}, \bibinfo{person}{Qiansheng Yang}, \bibinfo{person}{Jian Wang}, {and} \bibinfo{person}{Dongmei Fu}.} \bibinfo{year}{2022}\natexlab{}.
\newblock \showarticletitle{Dynamic Graph Reasoning for Multi-person 3D Pose Estimation}. In \bibinfo{booktitle}{\emph{Proceedings of the 30th ACM International Conference on Multimedia}}. \bibinfo{pages}{3521--3529}.
\newblock


\bibitem[Su et~al\mbox{.}(2022)]%
        {su2022virtualpose}
\bibfield{author}{\bibinfo{person}{Jiajun Su}, \bibinfo{person}{Chunyu Wang}, \bibinfo{person}{Xiaoxuan Ma}, \bibinfo{person}{Wenjun Zeng}, {and} \bibinfo{person}{Yizhou Wang}.} \bibinfo{year}{2022}\natexlab{}.
\newblock \showarticletitle{VirtualPose: Learning Generalizable 3D Human Pose Models from Virtual Data}. In \bibinfo{booktitle}{\emph{European Conference on Computer Vision}}. Springer, \bibinfo{pages}{55--71}.
\newblock


\bibitem[Tu et~al\mbox{.}(2020)]%
        {voxelpose}
\bibfield{author}{\bibinfo{person}{Hanyue Tu}, \bibinfo{person}{Chunyu Wang}, {and} \bibinfo{person}{Wenjun Zeng}.} \bibinfo{year}{2020}\natexlab{}.
\newblock \showarticletitle{VoxelPose: Towards Multi-Camera 3D Human Pose Estimation in Wild Environment}. In \bibinfo{booktitle}{\emph{European Conference on Computer Vision (ECCV)}}.
\newblock


\bibitem[Wang et~al\mbox{.}(2022)]%
        {Wang_2022_CVPR}
\bibfield{author}{\bibinfo{person}{Zitian Wang}, \bibinfo{person}{Xuecheng Nie}, \bibinfo{person}{Xiaochao Qu}, \bibinfo{person}{Yunpeng Chen}, {and} \bibinfo{person}{Si Liu}.} \bibinfo{year}{2022}\natexlab{}.
\newblock \showarticletitle{Distribution-Aware Single-Stage Models for Multi-Person 3D Pose Estimation}. In \bibinfo{booktitle}{\emph{Proceedings of the IEEE/CVF Conference on Computer Vision and Pattern Recognition (CVPR)}}. \bibinfo{pages}{13096--13105}.
\newblock


\bibitem[Wu et~al\mbox{.}(2021)]%
        {wu2021graph}
\bibfield{author}{\bibinfo{person}{Size Wu}, \bibinfo{person}{Sheng Jin}, \bibinfo{person}{Wentao Liu}, \bibinfo{person}{Lei Bai}, \bibinfo{person}{Chen Qian}, \bibinfo{person}{Dong Liu}, {and} \bibinfo{person}{Wanli Ouyang}.} \bibinfo{year}{2021}\natexlab{}.
\newblock \showarticletitle{Graph-based 3d multi-person pose estimation using multi-view images}. In \bibinfo{booktitle}{\emph{ICCV}}.
\newblock


\bibitem[Ye et~al\mbox{.}(2022)]%
        {fastervoxelpose}
\bibfield{author}{\bibinfo{person}{Hang Ye}, \bibinfo{person}{Wentao Zhu}, \bibinfo{person}{Chunyu Wang}, \bibinfo{person}{Rujie Wu}, {and} \bibinfo{person}{Yizhou Wang}.} \bibinfo{year}{2022}\natexlab{}.
\newblock \showarticletitle{Faster VoxelPose: Real-time 3D Human Pose Estimation by Orthographic Projection}. In \bibinfo{booktitle}{\emph{European Conference on Computer Vision (ECCV)}}.
\newblock


\bibitem[Zhang et~al\mbox{.}(2021)]%
        {zhang2021direct}
\bibfield{author}{\bibinfo{person}{Jianfeng Zhang}, \bibinfo{person}{Yujun Cai}, \bibinfo{person}{Shuicheng Yan}, \bibinfo{person}{Jiashi Feng}, {et~al\mbox{.}}} \bibinfo{year}{2021}\natexlab{}.
\newblock \showarticletitle{Direct multi-view multi-person 3d pose estimation}.
\newblock \bibinfo{journal}{\emph{Advances in Neural Information Processing Systems}}  \bibinfo{volume}{34} (\bibinfo{year}{2021}), \bibinfo{pages}{13153--13164}.
\newblock


\bibitem[Zhang et~al\mbox{.}(2020)]%
        {20204DAssociation}
\bibfield{author}{\bibinfo{person}{Yuxiang Zhang}, \bibinfo{person}{Liang An}, \bibinfo{person}{Tao Yu}, \bibinfo{person}{xiu Li}, \bibinfo{person}{Kun Li}, {and} \bibinfo{person}{Yebin Liu}.} \bibinfo{year}{2020}\natexlab{}.
\newblock \showarticletitle{4D Association Graph for Realtime Multi-person Motion Capture Using Multiple Video Cameras}. In \bibinfo{booktitle}{\emph{IEEE International Conference on Computer Vision and Pattern Recognition, (CVPR)}}.
\newblock


\bibitem[Zhen et~al\mbox{.}(2020)]%
        {zhen2020smap}
\bibfield{author}{\bibinfo{person}{Jianan Zhen}, \bibinfo{person}{Qi Fang}, \bibinfo{person}{Jiaming Sun}, \bibinfo{person}{Wentao Liu}, \bibinfo{person}{Wei Jiang}, \bibinfo{person}{Hujun Bao}, {and} \bibinfo{person}{Xiaowei Zhou}.} \bibinfo{year}{2020}\natexlab{}.
\newblock \showarticletitle{Smap: Single-shot multi-person absolute 3d pose estimation}. In \bibinfo{booktitle}{\emph{Computer Vision--ECCV 2020: 16th European Conference, Glasgow, UK, August 23--28, 2020, Proceedings, Part XV 16}}. Springer, \bibinfo{pages}{550--566}.
\newblock


\bibitem[Zhong et~al\mbox{.}(2018)]%
        {zhong2018camera}
\bibfield{author}{\bibinfo{person}{Zhun Zhong}, \bibinfo{person}{Liang Zheng}, \bibinfo{person}{Zhedong Zheng}, \bibinfo{person}{Shaozi Li}, {and} \bibinfo{person}{Yi Yang}.} \bibinfo{year}{2018}\natexlab{}.
\newblock \showarticletitle{Camera style adaptation for person re-identification}. In \bibinfo{booktitle}{\emph{Proceedings of the IEEE conference on computer vision and pattern recognition}}. \bibinfo{pages}{5157--5166}.
\newblock


\bibitem[Zhou et~al\mbox{.}(2022)]%
        {zhou2022quickpose}
\bibfield{author}{\bibinfo{person}{Zhize Zhou}, \bibinfo{person}{Qing Shuai}, \bibinfo{person}{Yize Wang}, \bibinfo{person}{Qi Fang}, \bibinfo{person}{Xiaopeng Ji}, \bibinfo{person}{Fashuai Li}, \bibinfo{person}{Hujun Bao}, {and} \bibinfo{person}{Xiaowei Zhou}.} \bibinfo{year}{2022}\natexlab{}.
\newblock \showarticletitle{QuickPose: Real-time Multi-view Multi-person Pose Estimation in Crowded Scenes}. In \bibinfo{booktitle}{\emph{ACM SIGGRAPH 2022 Conference Proceedings}}. \bibinfo{pages}{1--9}.
\newblock


\end{thebibliography}

\end{document}